\documentclass[a4paper,10pt]{article}
\usepackage[utf8]{inputenc}
\usepackage{amsmath,bbm} 
\usepackage{amsfonts} 
\usepackage{amssymb}
\usepackage{theorem}
\usepackage{pifont}
\usepackage{mathabx}
\usepackage{url}
\usepackage{hyperref}
\usepackage{graphicx}

\def\P{{\mathbb{P}}}
\def\EE{{\mathbb{E}}}

\newtheorem{defi}{Definition}
\newtheorem{prop}{Proposition}

\newcommand{\finproof}{ 
\begin{flushright}
\ding{110}
\end{flushright}
}
\newenvironment{proof}{
\begin{flushleft}
\textbf{Proof:}
\end{flushleft}
}{
\finproof
}
\title{A parameterless scale-space approach to find meaningful modes in histograms - Application to image and spectrum segmentation}
\author{J\'er\^ome Gilles, Kathryn Heal}

\graphicspath{{./Figures/}}

\begin{document}

\maketitle

\section{Introduction}
Despite a huge literature on this topic, image segmentation remains a difficult problem in the sense there does not exist a general 
method which works in all cases. One reason is in that the expected segmentation generally depends on the final application goal. 
Generally researchers focus on the development of specific algorithms according to the type of images they are processing and their 
final purpose (image understanding, object detection, \ldots). Different types of approaches were developed in the past which can 
be broadly classified into histogram based, edge based, region based and clustering (and mixes between them). If histogram methods 
are conceptually straightforward, they are not the most efficient but they are still widely used because of their simplicity 
and the few computational resources needed 
to perform them (this is an important criterion for computer vision applications). The idea behind such methods is that the final 
classes in the segmented image correspond to ``meaningful'' modes in an histogram built from the image characteristics. For instance,
 in the case of grayscale images, each class is supposed to correspond to a mode in the histogram of the gray values. Finding 
such modes is basically equivalent to find a set of thresholds separating the mode supports in the histogram. Several articles are 
available in the literature proposing histogram based segmentation algorithms. Two main philosophies can be encountered: techniques 
using histograms to drive a more advanced segmentation algorithm or techniques based on the segmentation of the histogram itself.
For instance, we can cite the work of Chan et al. \cite{Chan2007}, where the authors compare the empirical histograms of 
two regions (binary segmentation) by using the Wasserstein distance in a levelset formulation.
In \cite{Yuan2012}, local spectral histograms (i.e obtained by using several filters) are built. The authors show that the 
segmentation process is equivalent to solving a linear regression problem. Based on their formalism, they also propose a method to 
estimate the number of classes.
In \cite{Puzicha1999}, a mixture model for histogram data is proposed and then used to perform the final clustering (the number of 
clusters is chosen accordingly with some rules from the statistical learning theory).
In \cite{Sural2002}, it is shown that the HSV color space is a better color representation space than the usual RGB space. The 
authors use $k-$Means to obtain the segmentation. They also show that this space can be used to build feature histograms to perform an 
image retrieval task.
Another approach, called JND (Just Noticeable Difference) histogram, is used in \cite{Bhoyar2010} to build a single histogram 
which describes the range of colors. This type of histogram is based on the human perception capabilities. The authors propose a simple 
algorithm to segment the JND histogram and get the final classes. This method still has some parameters and therefore we do not see it as being optimal.
In \cite{Kurugollu}, the authors build 2D histograms from the pairs RB-RG-GB (from the RGB cube), then segment them by assigning each 
histogram pixel to their most attractive closest peak. The attraction force is based on distances to each histogram peak and their 
weights. This last step consists in the fusion of each segmentation to form a global one.
In \cite{Yildizoglu2013}, a variational model embedding histograms is proposed to segment an image where some reference histograms are 
supposed to be known.
An interesting approach was investigated in \cite{Delon2007}. The authors propose a fully automatic algorithm to detect the modes in 
an histogram $H$. It is based on a fine to 
coarse segmentation of $H$. The algorithm is initialized with all local minima of $H$. The authors defined a statistical criteria, 
based on the Grenander estimator, to 
decide if two consecutive supports correspond to a common global trend or if they are parts of true separated modes; this criteria is 
based on the $\epsilon-$meaningful events theory \cite{Delon2007,Desolneux2008}. The (parameterless) algorithm can be resumed as 
following: start from the finest segmentation given by all local minima, choose one minimum and check if adjacent supports are part of 
a same trend or not. If yes then merge these supports by removing this local minima from the list. Repeat until no merging are 
possible. This work is extended to color images in \cite{Delon2007a} where they successively apply the 
previous algorithm on the different components H,S,V and use this segmentation result to initialize a $k-$Means algorithm to get the 
final segmentation. While this approach provides a fully automatic algorithm, it becomes computationally expensive (both in terms of 
time and memory) for histograms defined on a large set of bins.\\
Another application of such histogram modes detection is the identification of ``harmonic'' modes in Fourier spectra (indeed a 
spectrum can be seen as an histogram counting the occurrence of each frequency). For instance, the ability to find the set of supports 
of such modes is fundamental to build the new empirical wavelets proposed in \cite{ewt1d,ewt2d}. It is to notice that working with 
spectra can be more difficult in the sense that they, generally, are less regular than classic histograms and that it is difficult 
to have an a priori idea of the needed relevant number of modes.\\
In this paper, we propose an algorithm to automatically find meaningful modes in an histogram or spectrum. Our approach is based on a scale-space 
representation of the considered histogram which permits us to define the notion of ``meaningful modes'' in a simpler way. We will 
show that finding $N$ (where $N$ is itself unknown) modes is equivalent to perform a two class clustering. This method is simple and 
runs very fast. The remainder of the paper is organized as follows: in section~\ref{sec:ss}, we recall the definition and properties 
of a scale-space representation and build our model. In section~\ref{sec:expe}, we present several experiments in histogram 
segmentation, grayscale image segmentation and color reduction as well as the detection of modes in a signal spectrum. Finally, we draw 
conclusions in section~\ref{sec:conc}.

\section{Scale-space histogram segmentation} \label{sec:ss}
\subsection{Scale-space representation}
Let a function $f(x)$ be defined over an interval $[0,x_{max}]$ and let the kernel $g(x;t)=\frac{1}{\sqrt{2\pi t}}e^{-x^2/(2t)}$. 
The scale-space representation \cite{Witkin1984} of $f$ is given by 
($\otimes$ denotes the convolution product)
\begin{equation}
L(x,t)=g(x;t)\otimes f(x)
\end{equation}
This operation removes all ``patterns'' of characteristic length $\sqrt{t}$ i.e. as $t$ increases, $L(x,t)$ becomes smoother. 
This operator fulfill a semi-group property: $g(x;t_1+t_2)=g(x;t_1)\otimes g(x;t_2)$.
This means that we can iterate the convolution to get $L(x,t)$ at different scales (this is not valid for a discretized version of the Gaussian 
kernel except if the ratio $t_2/t_1$ is odd). We choose to start with $\sqrt{t_0}=0.5$ because since we work with finite length signals, there is no 
interest to go further than $\sqrt{t_{max}}=x_{max}$. In practice, we want to perform a finite number of steps, denoted $N_{step}$, 
to go from the initial scale to the final one. Thus we can write: $\sqrt{t_{max}}=N_{step}\sqrt{t_0}$ which implies 
$N_{step}=2x_{max}$.\\
In this paper, we chose to use the sampled Gaussian kernel to implement the scale-space representation:
\begin{equation}
L(x,t)=\sum_{n=-\infty}^{+\infty}f(x-n)g(n;t),
\end{equation}
where
\begin{equation}
g(n;t)=\frac{1}{\sqrt{2\pi t}}e^{-n^2/2t}.
\end{equation}
In practice we use a truncated filter in order to have a finite impulse response filter:
\begin{equation}
L(x,t)=\sum_{n=-M}^{+M}f(x-n)g(n;t),
\end{equation}
with $M$ large enough that the approximation error of the Gaussian is negligible. A common choice 
is to set $M=C\sqrt{t}+1$ with $3\leq C\leq 6$ (this means that the filter's size is increasing with respect to $t$). 
In our experiments we fix $C=6$ in order to ensure an approximation error 
smaller than $10^{-9}$.

\subsection{Meaningful scale-space modes}
\begin{figure}[!h]
\begin{center}
\begin{tabular}{cc}
\includegraphics[scale=0.255]{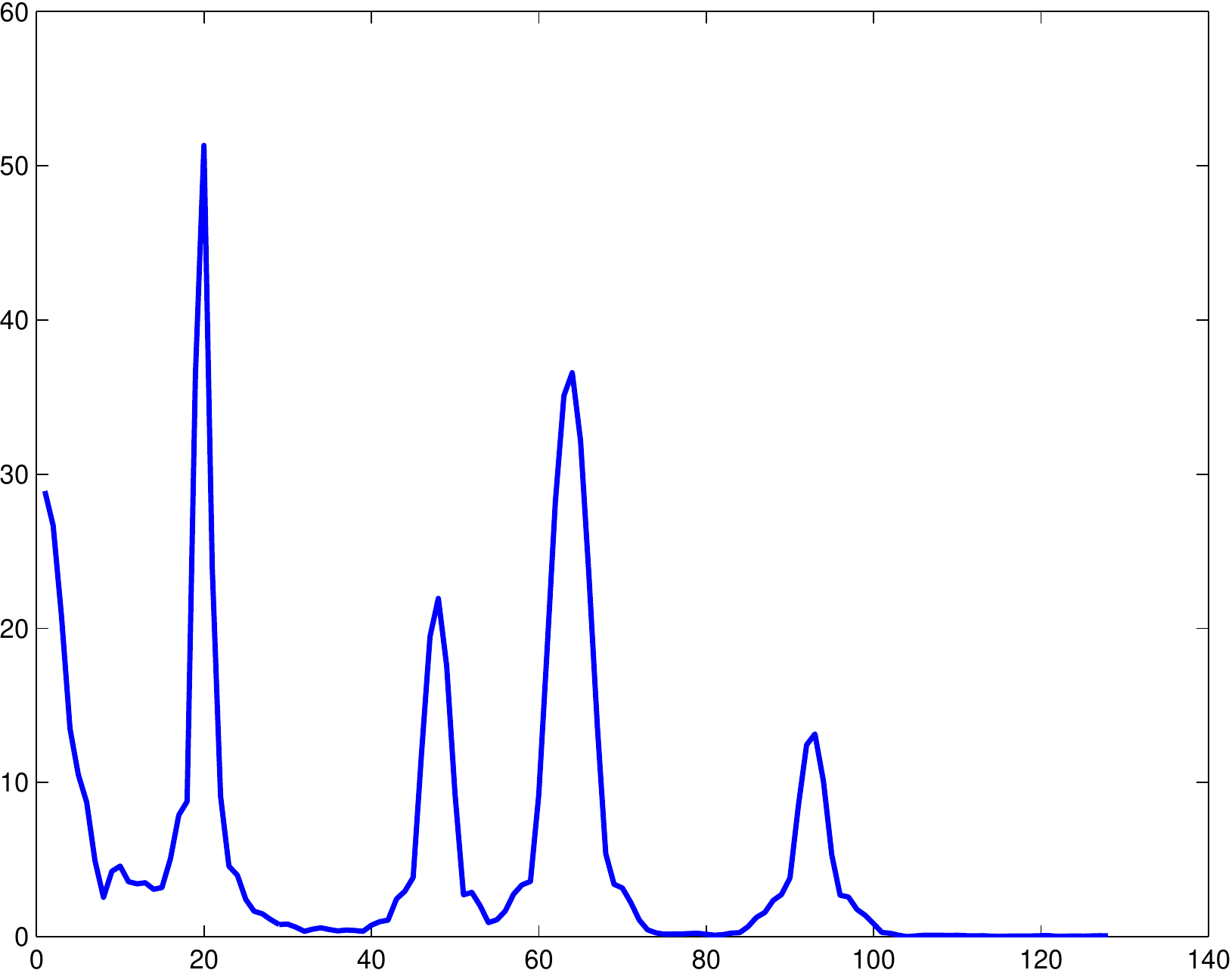} &
\fbox{\includegraphics[scale=0.7]{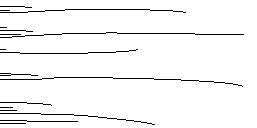}}\\
(a) & (b)
\end{tabular}
\end{center}
\caption{Example of modes in an histogram (a) and its corresponding scale-space representation (b) where the horizontal axis 
corresponds to the scale $t$ and the vertical one to the original horizontal axis of (a), respectively.}
\label{fig:modes}
\end{figure}

Our objective is to find meaningful modes in a given histogram; hence we first need to define the notion of ``meaningful mode''. 
We will begin by defining what is a mode, and explaining its representation in the scale-space plane. Let us consider an histogram like the 
one depicted in figure~\ref{fig:modes}.a. It is clear that to find boundaries delimiting consecutive modes is essentially to find 
intermediate valleys or equivalently local minima in the histogram. 
In order to be able to define the notion of ``meaningful'' modes, we will use one of the most important properties of scale-space 
representations (this notion was already used by the computer vision community to detect edges in an image): the number of minima 
with respect to $x$ of $L(x,t)$
is a decreasing function of the scale parameter $t$ and no new minima can appear as $t$ increases. For instance, figure~\ref{fig:modes}.b provides 
the scale-space plane corresponding to the figure~\ref{fig:modes}.a. Observe that each of the initial minima (for $t=0$) generates a curve 
in the scale-space plane. Let us fix some notations. The number of initial minima is denoted $n$, and each of the local minima defines a 
``scale-space curve'' $C_i$ ($i\in[1,n]$) of length $L_i$. We can now define the notion of ``meaningful modes'' of an histogram.

\begin{defi}
A mode in an histogram is called meaningful if its support is delimited by consistent, with respect to $t$, 
local minima i.e. minima which generate long scale-space curves $C_i$.
\end{defi}
As a consequence, finding meaningful modes is equivalent to find a threshold $T$ such that scale-space curves of length larger than 
$T$ are the curves corresponding to consistent minima. This means that the problem of finding such modes is equivalent to a two 
class clustering problem on the set $\{L_i\}_{i\in[1,n]}$. The following sections explore different ways to automatically 
find the expected threshold $T$.

\subsection{Probabilist approach}
In detection problems it is typical to use probabilistic models. We will investigate several possible distribution laws but first we need to 
translate the notion of meaningful mode in terms of probability.
\begin{defi}
Given a positive small number $\epsilon$, a minimum $i$ will to be said $\epsilon$-meaningful if
\begin{equation}\label{eq:test}
\P(L_i>T)\leq\epsilon.
\end{equation} 
\end{defi}
Based on this definition, the following proposition gives explicit expressions of $T$ for different distribution laws.
\begin{prop}
Assuming the variables $L_i$ are independent random variables (where $1\leq L_i\leq L_{max}$) and $\epsilon$  is a positive small 
number, if we denote $H_L$ the histogram representing the occurrences of the lengths of a scale-space curve, we have
\begin{itemize}
\item if $\P$ is the uniform distribution: $T\geq (1-\epsilon)L_{max}+1$,
\item if $\P$ is the half-normal distribution: $T\geq\sqrt{2\sigma^2}erf^{-1}\left(erf\left(\frac{L_{max}}{\sqrt{2\sigma^2}}\right)-\epsilon\right)$ 
(where accordingly to the definition of the half-normal distribution, $\sigma=\sqrt{\frac{\pi}{2}}\EE[H_L]$),
\item if $\P$ is the empirical distribution: $T$ is empirically chosen such that $$\sum_{k=1}^TH_L(k)=(1-\epsilon)\sum_{k=1}^{L_{max}}H_L(k).$$
\end{itemize}
\end{prop}
Before we give the proof of this proposition, let us comment the choice of these distribution laws. The uniform distribution is the 
law which will not privilege any range for $T$. In many practical cases, a Gaussian law is chosen as a default distribution law but 
in our context, we know that the random variables $L_i$ are always positives and in many experiments they follow a monotone decreasing 
law hence our choice of the half-normal law \cite{Azzalini}. The empirical distribution is used when no a priori information is known and the 
entire model is based on empirical measurements. Note that in practice a simple choice for $\epsilon$ is to take $\epsilon=1/n$. 

\begin{proof}
We first consider the uniform law then
\begin{align}
\P(L_i>T)&=\sum_{k=T}^{L_{max}}\P(L_i=k)\\
&=\sum_{k=T}^{L_{max}}\frac{1}{L_{max}}\\
&=\frac{L_{max}+1-T}{L_{max}}
\end{align}
then \eqref{eq:test} gives
\begin{equation}
\frac{L_{max}+1-T}{L_{max}}\leq\epsilon\Leftrightarrow T\geq (1-\epsilon)L_{max}+1.
\end{equation}
Next we address the half-normal law, we have
\begin{align}
\P(L_i>T)&=\int_T^{L_{max}}\P(L_i=x)dx\\
&=\int_T^{L_{max}}\sqrt{\frac{2}{\pi\sigma^2}}\exp\left(-\frac{x^2}{2\sigma^2}\right)dx\\
&=\left[\int_0^{L_{max}}\sqrt{\frac{2}{\pi\sigma^2}}\exp\left(-\frac{x^2}{2\sigma^2}\right)dx-\int_0^T\sqrt{\frac{2}{\pi\sigma^2}}\exp\left(-\frac{x^2}{2\sigma^2}\right)dx\right]\\
&=erf\left(\frac{L_{max}}{\sqrt{2\sigma^2}}\right)-erf\left(\frac{T}{\sqrt{2\sigma^2}}\right)
\end{align}
then \eqref{eq:test} gives
\begin{eqnarray}
erf\left(\frac{L_{max}}{\sqrt{2\sigma^2}}\right)-erf\left(\frac{T}{\sqrt{2\sigma^2}}\right)\leq\epsilon\\
\Leftrightarrow T\geq\sqrt{2\sigma^2}erf^{-1}\left(erf\left(\frac{L_{max}}{\sqrt{2\sigma^2}}\right)-\epsilon\right).
\end{eqnarray}
Finally, we consider the empirical case. We use the histogram, denoted $H_L$, of the lengths $L_i$. 
Basically, $H_L(k)$ is the number of curves having a length $k$ (because we build the scale-space representation 
over a finite set of scales, we know that there is a maximal length $L_{max}$, i.e. $k\in [1,L_{max}]$). 
Then we keep the threshold $T$ such that $\sum_{k=1}^TH_L(k)=(1-\epsilon)\sum_{k=1}^{L_{max}}H_L(k)$.
This conclude the proof.
\end{proof}

\subsection{Otsu's method}
In \cite{Otsu}, the author proposed an algorithm to separate an histogram $H_L$ (defined as in the previous section) into two classes 
$H_1$ and $H_2$. The goal of Otsu's method is to find the threshold $T$ such that the intra-variances of each class $H_1,H_2$ are 
minimal and the inter class variance 
is maximal. This corresponds to finding $T$ (an exhaustive search is done in practice) which maximizes the between class variance $\sigma_B^2=W_1W_2(\mu_1-\mu_2)^2$, where 
$W_r=\frac{1}{n}\sum_{k\in H_r}H(k)$ and $\mu_r=\frac{1}{n}\sum_{k\in H_r}kH(k)$.

\subsection{$k-$Means}
The aim of the $k-$Means algorithm is to partition a set of points into $k$ clusters. In our context, we apply the $k-$Means 
algorithm to the histogram $H_L$ to get the two clusters 
$H_1,H_2$ (meaningful/non-meaningful minima). It corresponds to the following problem:
\begin{equation}\label{eq:kmeans}
(H_1,H_2)=\underset{H_1,H_2}{\arg\min} \sum_{r=1}^2\sum_{H(k)\in H_r}\|H(k)-\mu_r\|^2,
\end{equation}
where $\mu_r$ is the mean of points in $H_r$. In this paper, we experiment with both the $\ell^1$ and $\ell^2$ norms and two types of 
initialization: random or uniformly distributed. In practice, it is usual to 
run the $k-$Means several times and keep the solution that provides the smallest minimum of \eqref{eq:kmeans} 
(in our experiments we chose to perform ten iterations).


\section{Experiments}\label{sec:expe}
\subsection{1D histogram segmentations}
\begin{figure}[!h]
\begin{center}
\begin{tabular}{cc}
\includegraphics[width=0.47\textwidth]{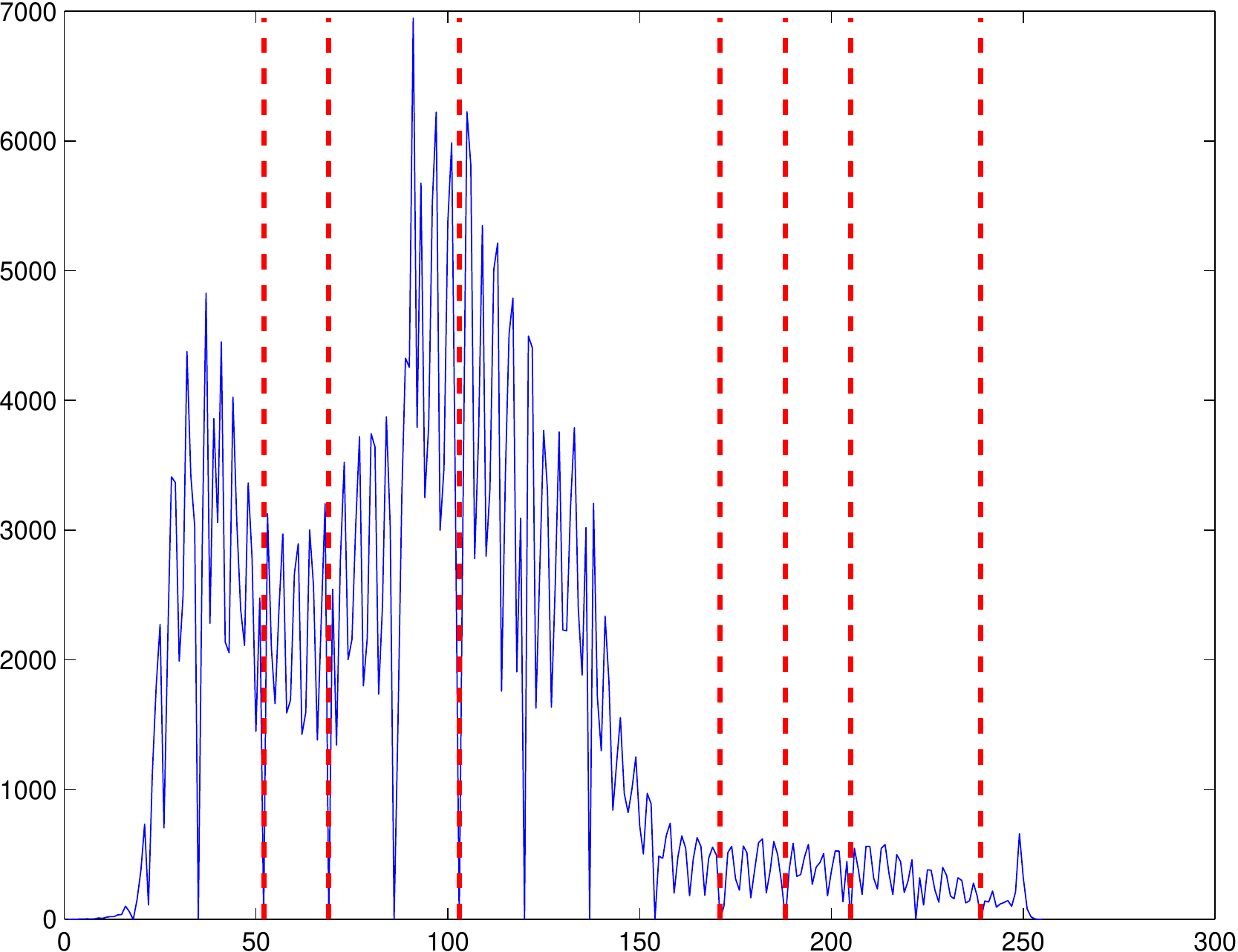} &
\includegraphics[width=0.47\textwidth]{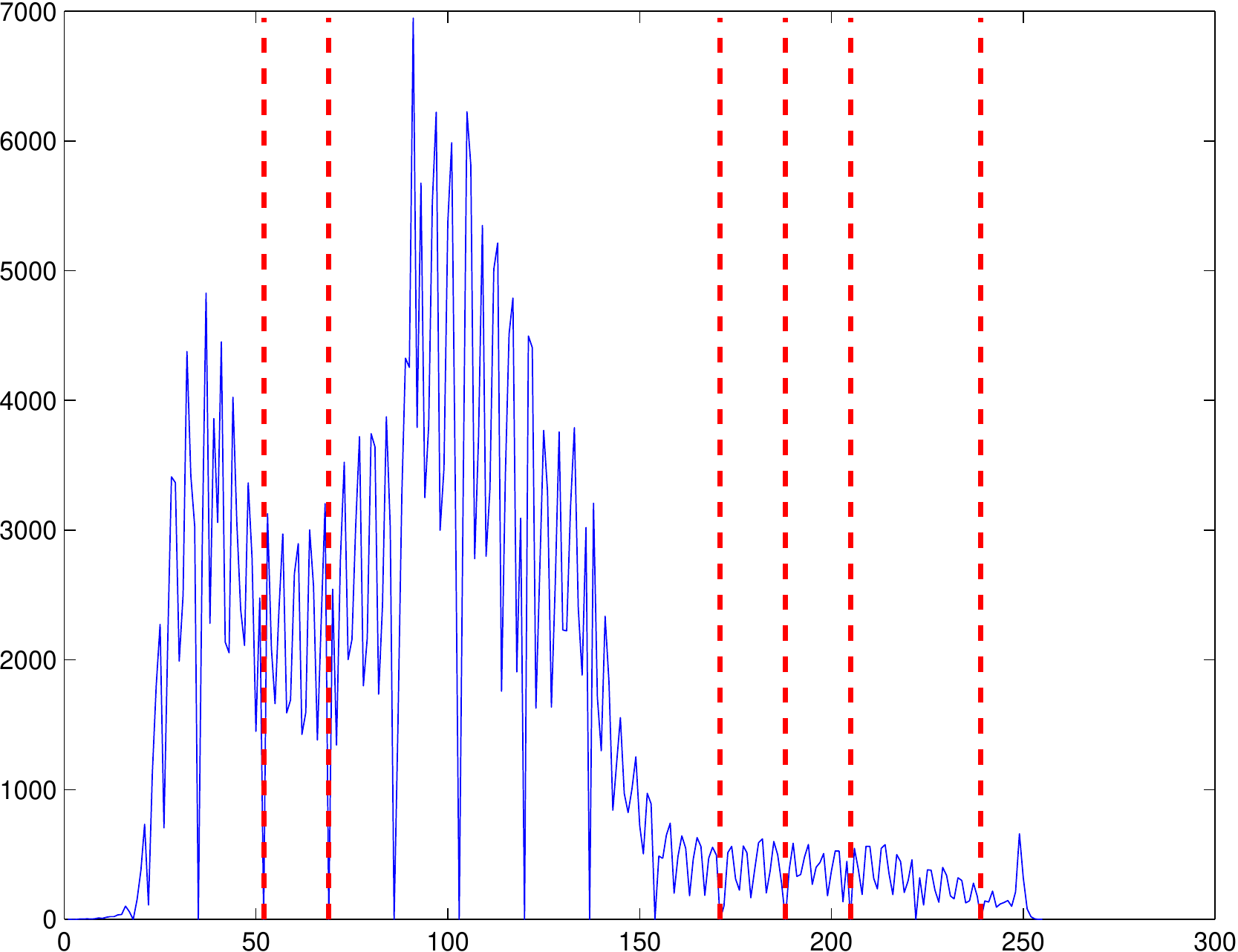} \\
Otsu & $k-$Means and Half-Normal law
\end{tabular}
\caption{Boundaries for $x16$.}
\label{fig:hx16}
\end{center}
\end{figure}

\begin{figure}[!h]
\begin{center}
\begin{tabular}{cc}
\includegraphics[width=0.47\textwidth]{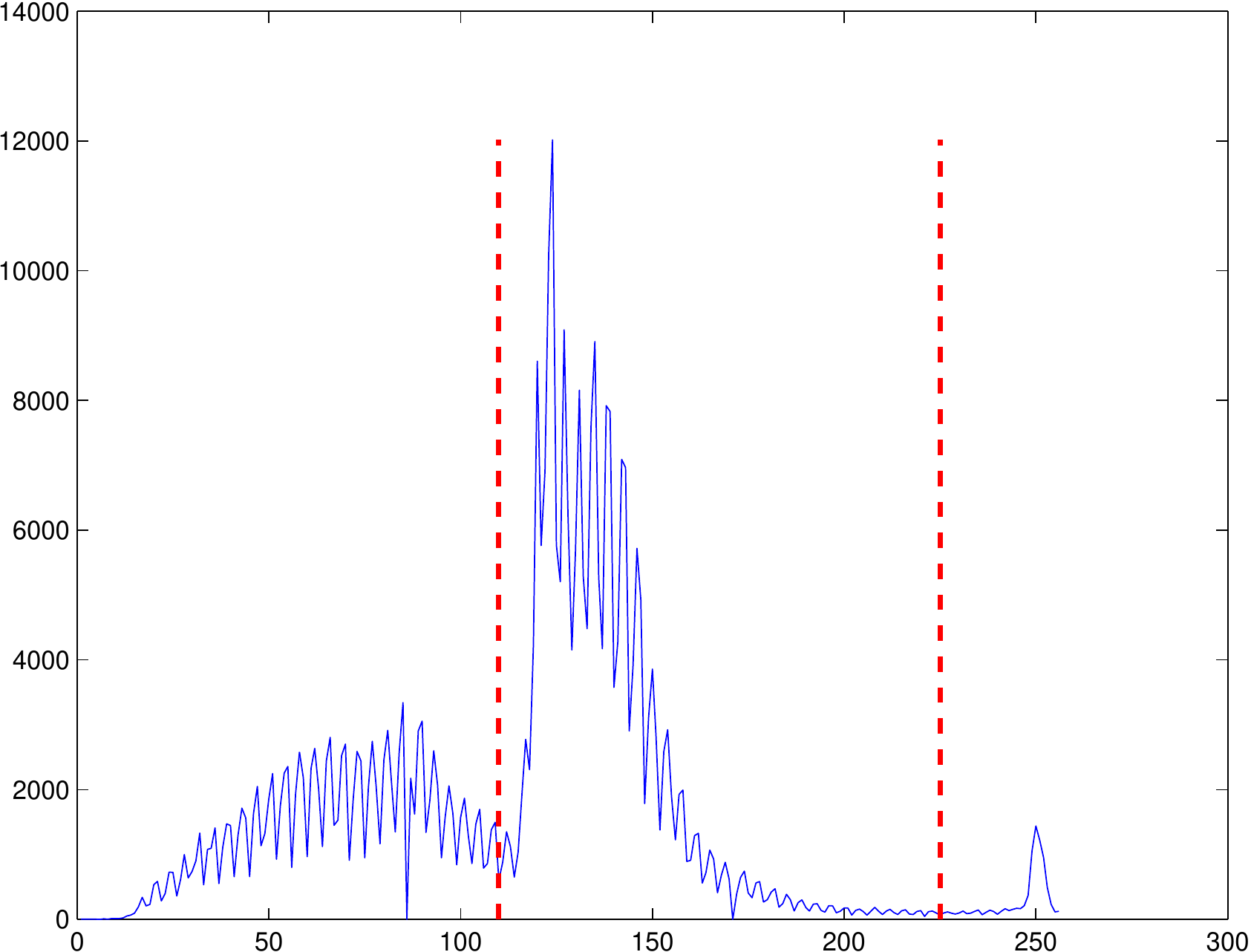} &
\includegraphics[width=0.47\textwidth]{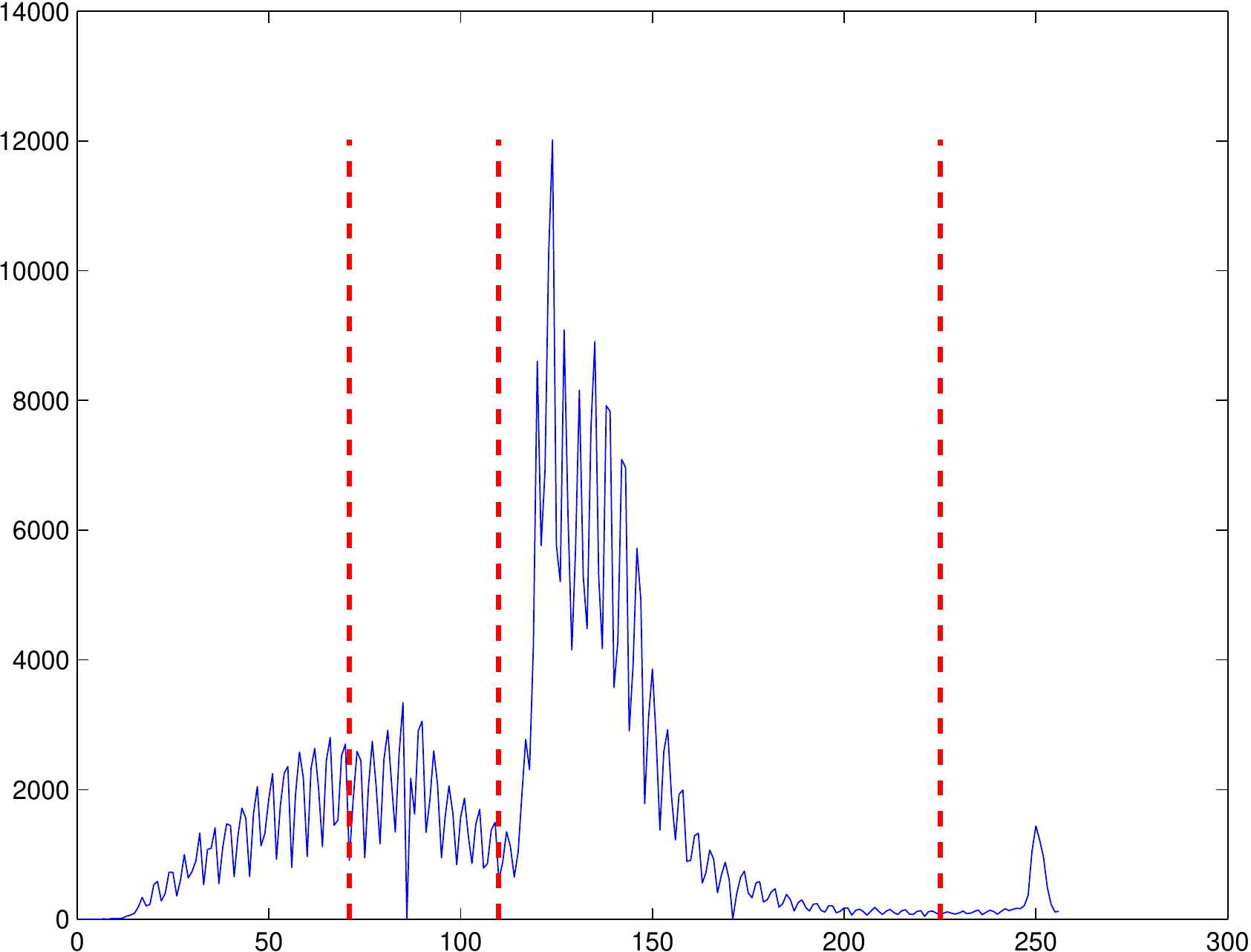} \\
$k-$Means and Empirical law & Otsu and Half-Normal law
\end{tabular}
\caption{Boundaries for $x21$.}
\label{fig:hx21}
\end{center}
\end{figure}

\begin{figure}[!h]
\begin{center}
\begin{tabular}{cc}
\includegraphics[width=0.47\textwidth]{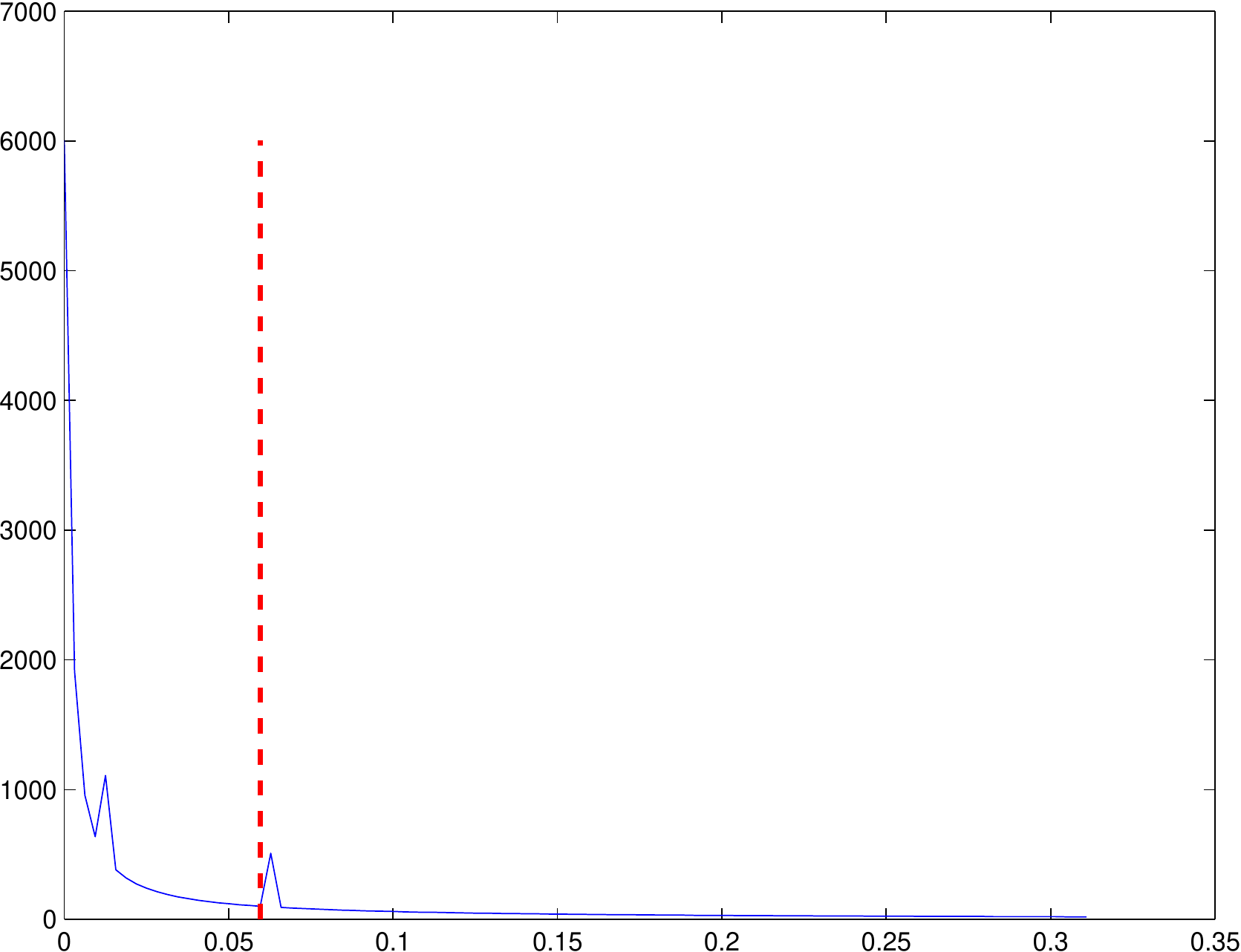} &
\includegraphics[width=0.47\textwidth]{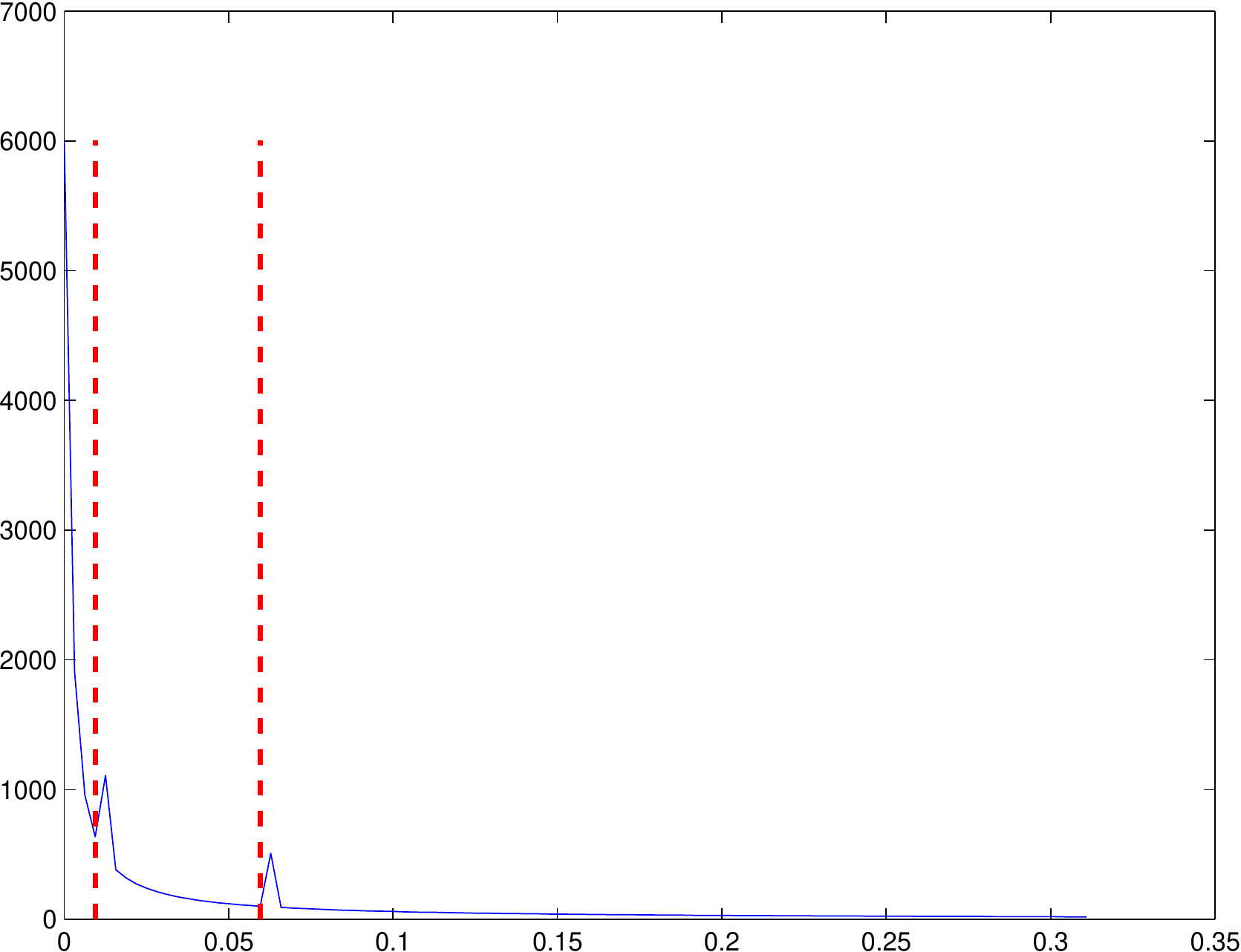} \\
$k-$Means and Empirical law & Otsu and Half-Normal law
\end{tabular}
\caption{Boundaries for $Sig1$.}
\label{fig:bsig1}
\end{center}
\end{figure}

\begin{figure}[!h]
\begin{center}
\begin{tabular}{cc}
\includegraphics[width=0.47\textwidth]{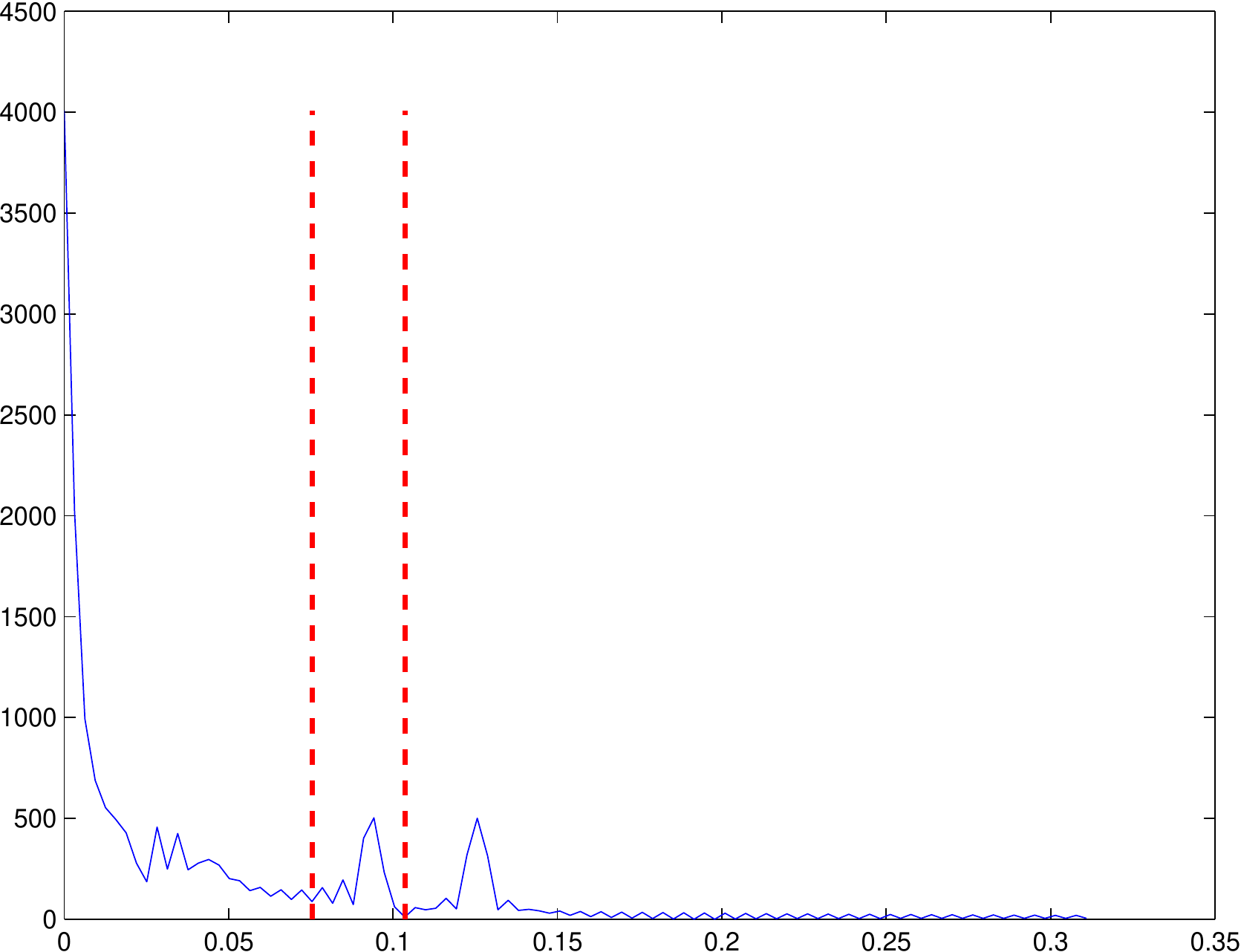} &
\includegraphics[width=0.47\textwidth]{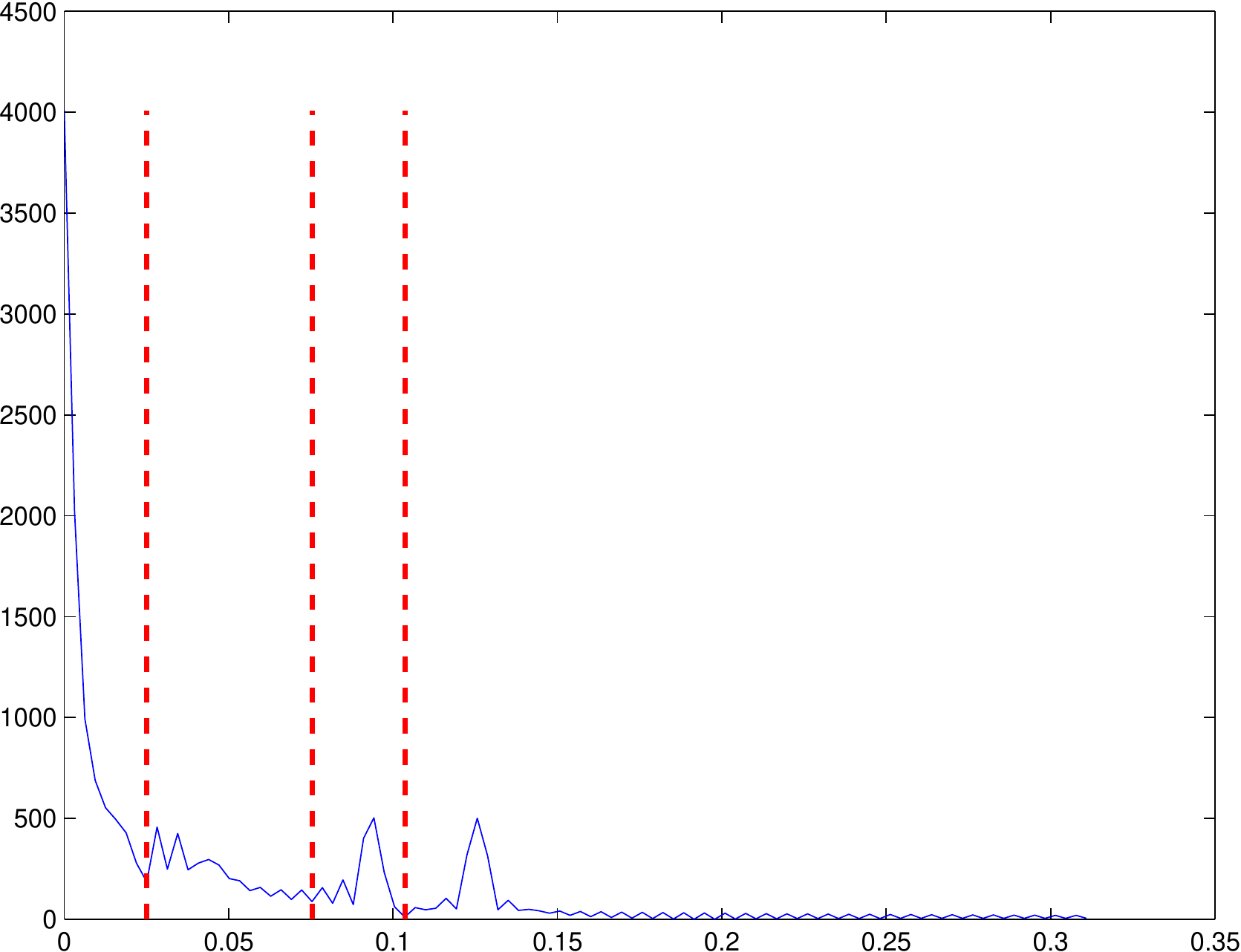} \\
$k-$Means and Empirical law & Otsu and Half-Normal law
\end{tabular}
\caption{Boundaries for $Sig2$.}
\label{fig:bsig2}
\end{center}
\end{figure}

\begin{figure}[!h]
\begin{center}
\begin{tabular}{cc}
\includegraphics[width=0.47\textwidth]{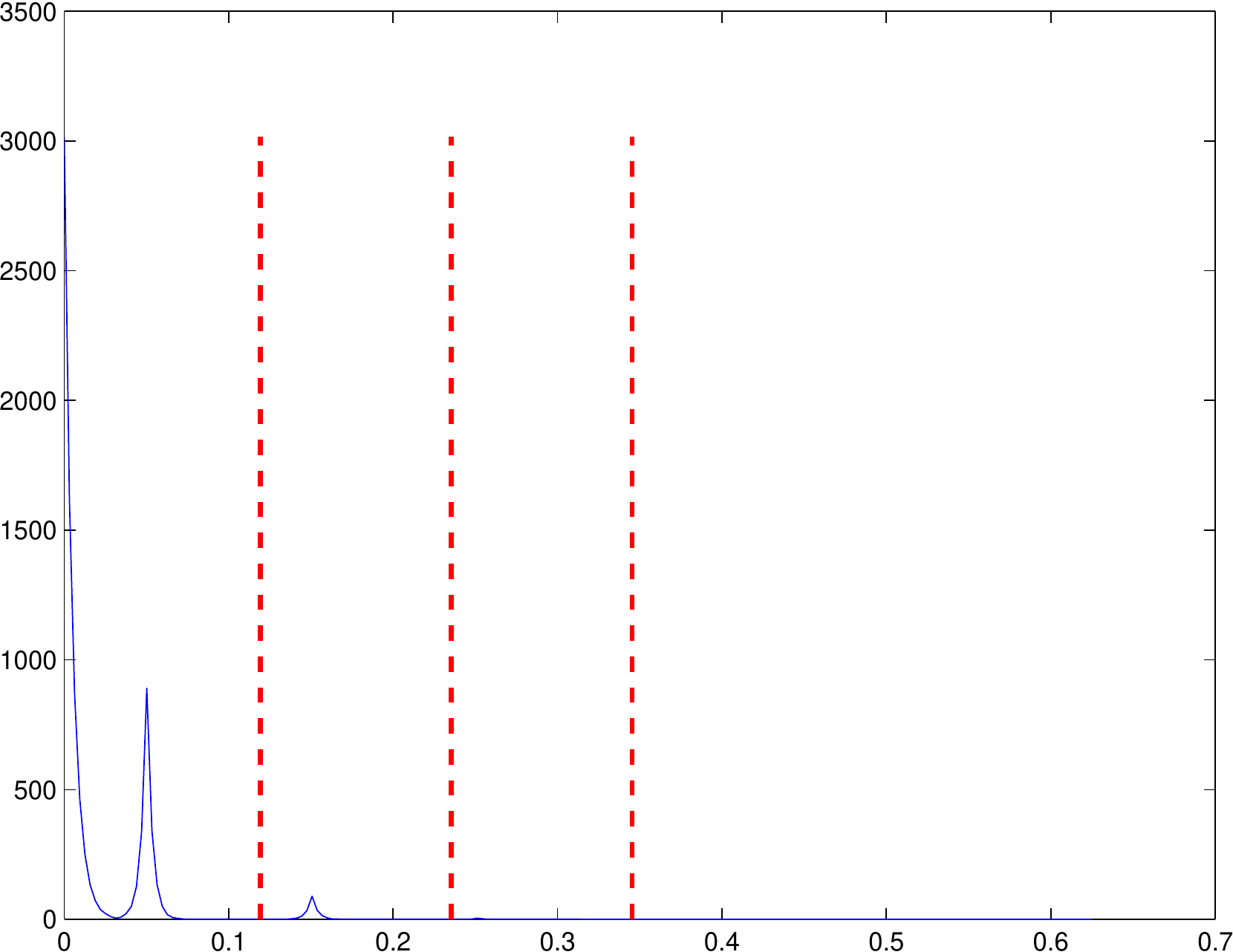} &
\includegraphics[width=0.47\textwidth]{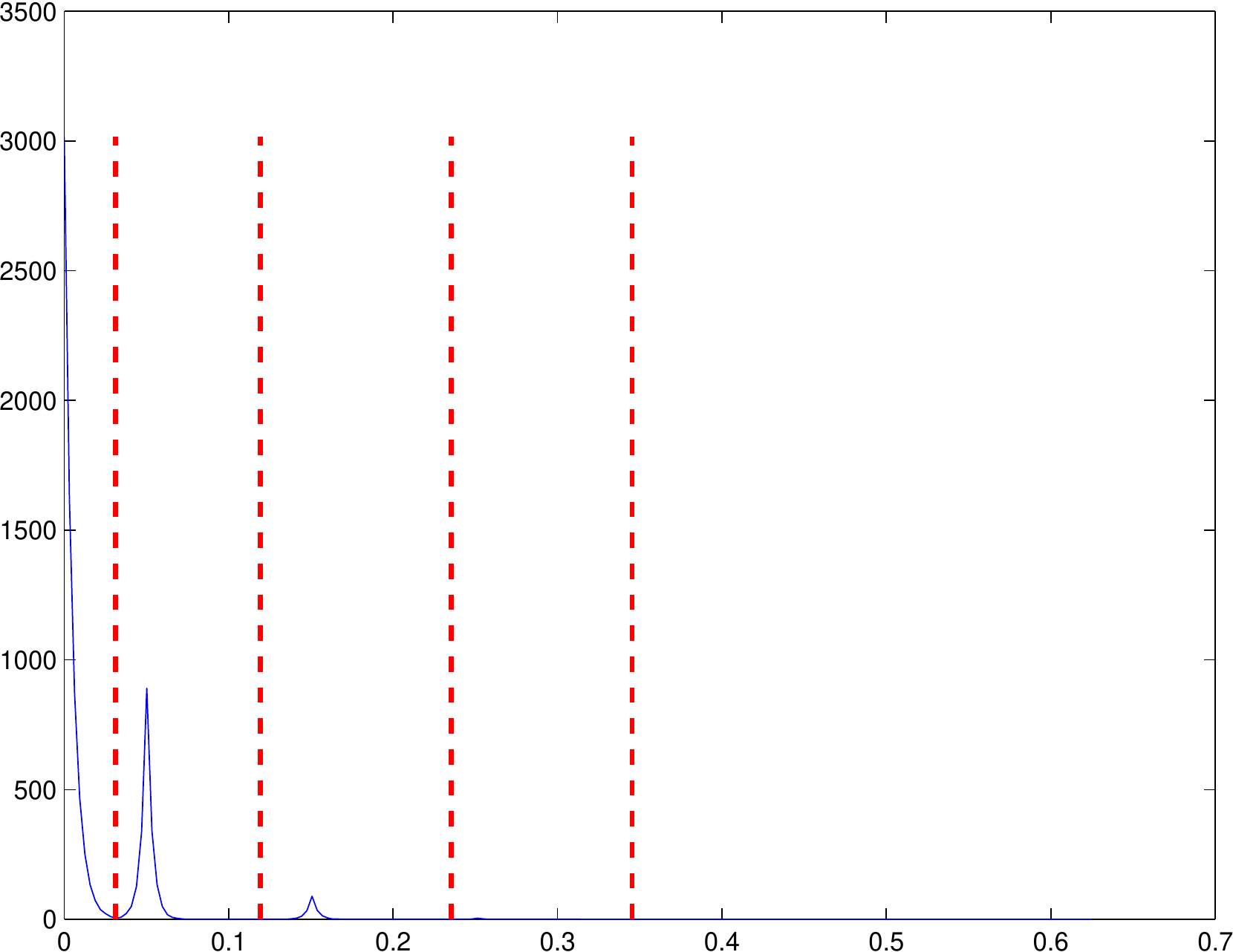} \\
$k-$Means & Otsu and Half-Normal law
\end{tabular}
\caption{Boundaries for $Sig3$ (results from the Empirical law are not represented as not interesting).}
\label{fig:bsig3}
\end{center}
\end{figure}

\begin{figure}[!h]
\begin{center}
\begin{tabular}{cc}
\includegraphics[width=0.47\textwidth]{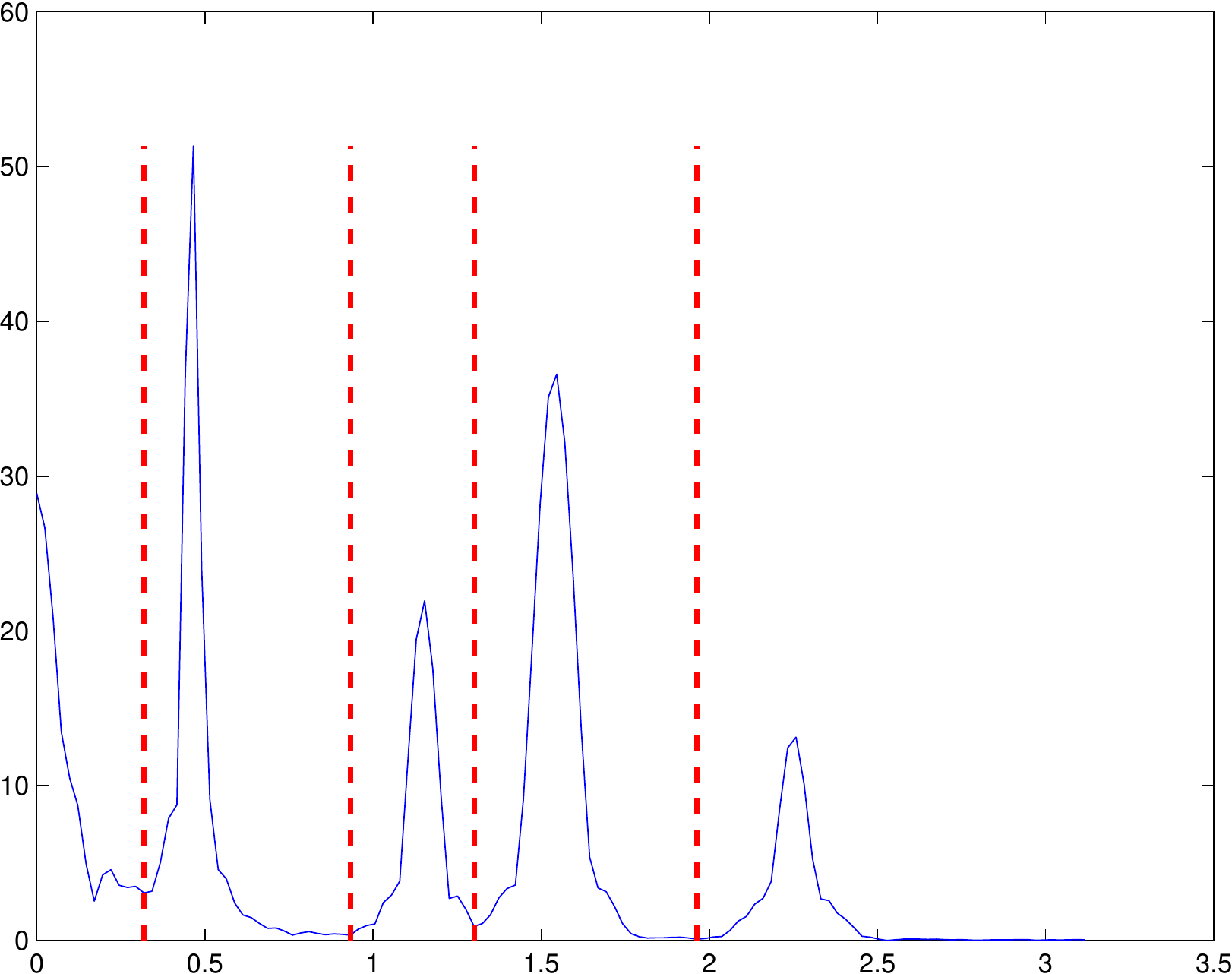} &
\includegraphics[width=0.47\textwidth]{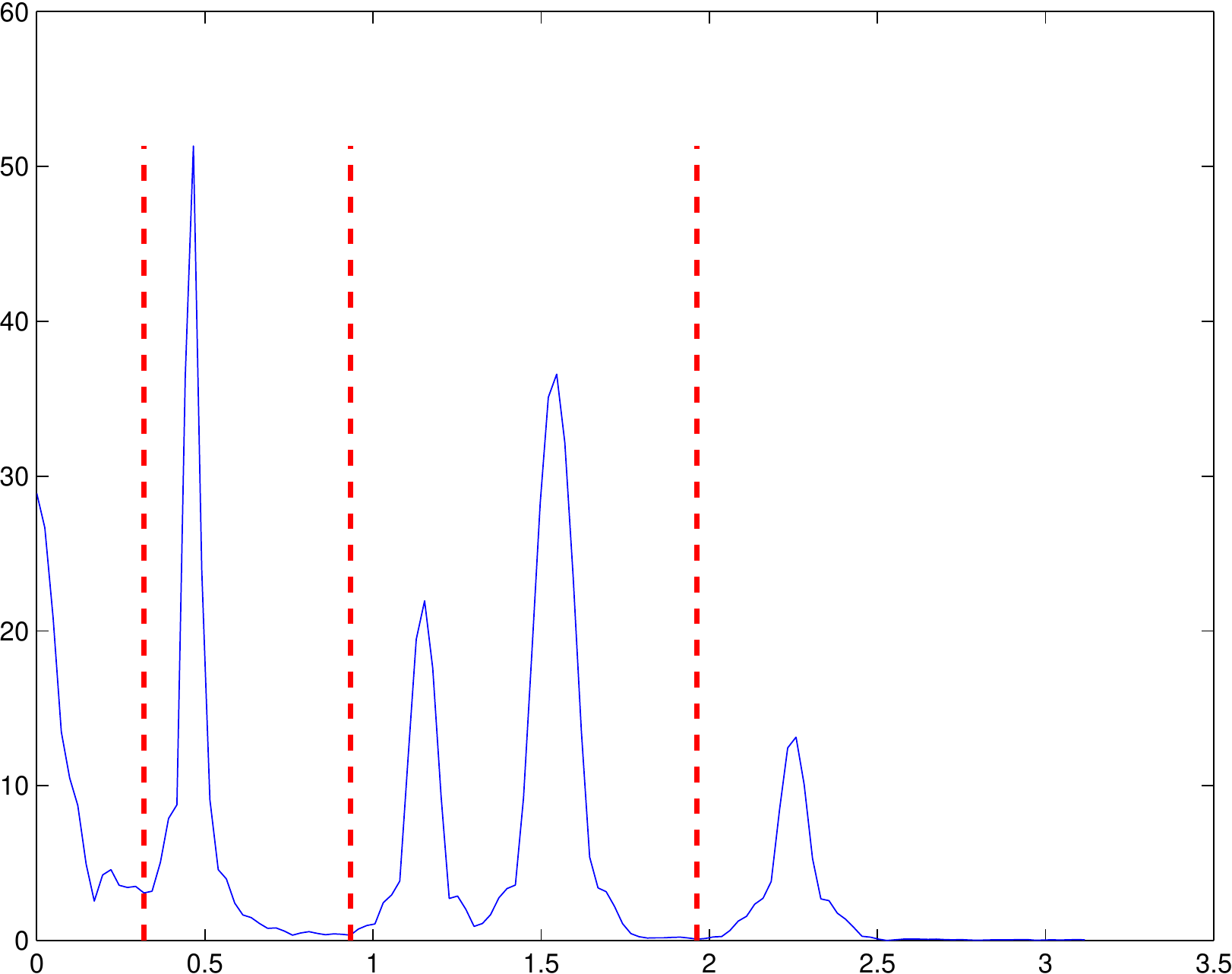} \\
$\ell^2-k-$Means & $\ell^1-k-$Means \\
\includegraphics[width=0.47\textwidth]{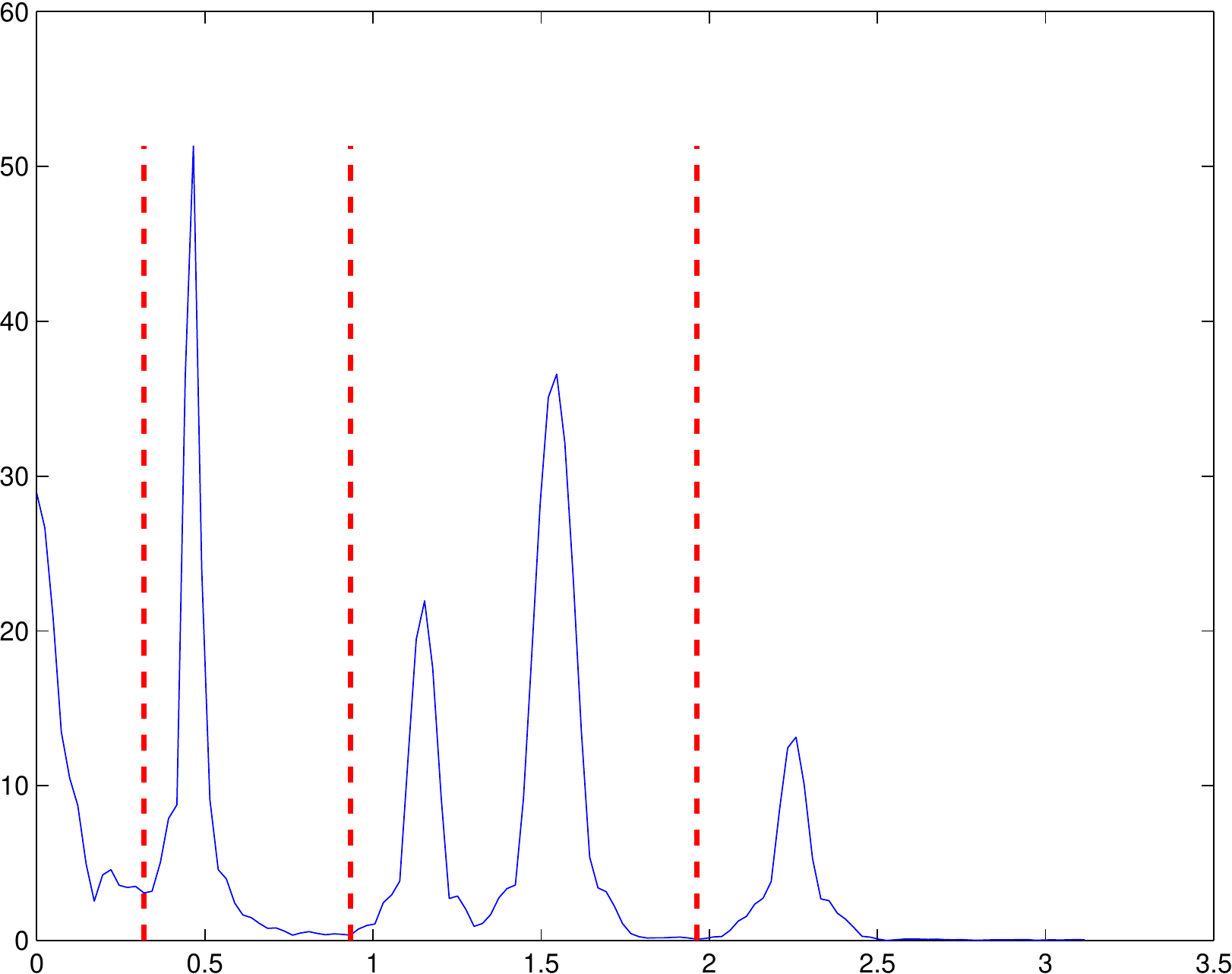} &
\includegraphics[width=0.47\textwidth]{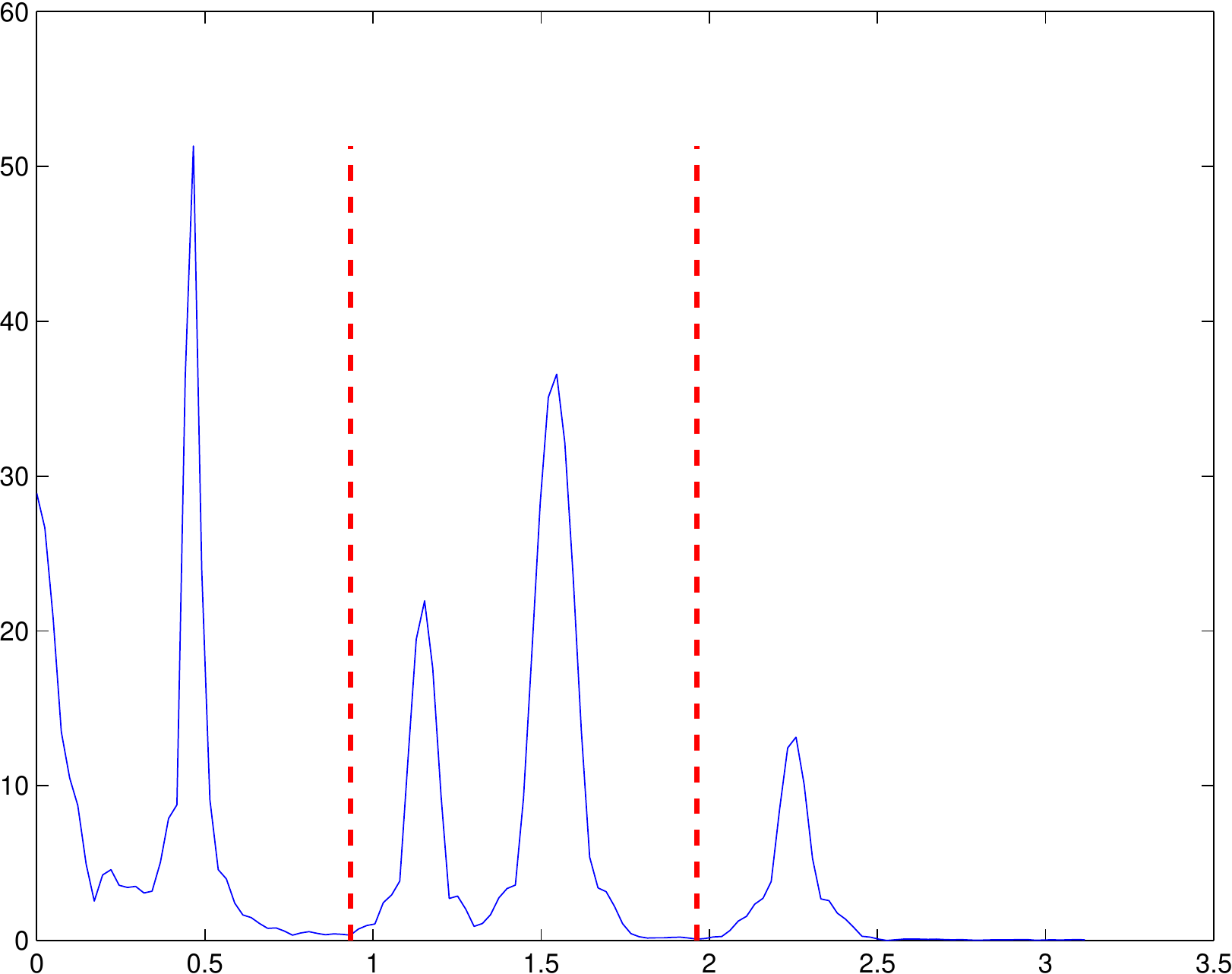} \\
Half-Normal & Empirical law \\
\includegraphics[width=0.47\textwidth]{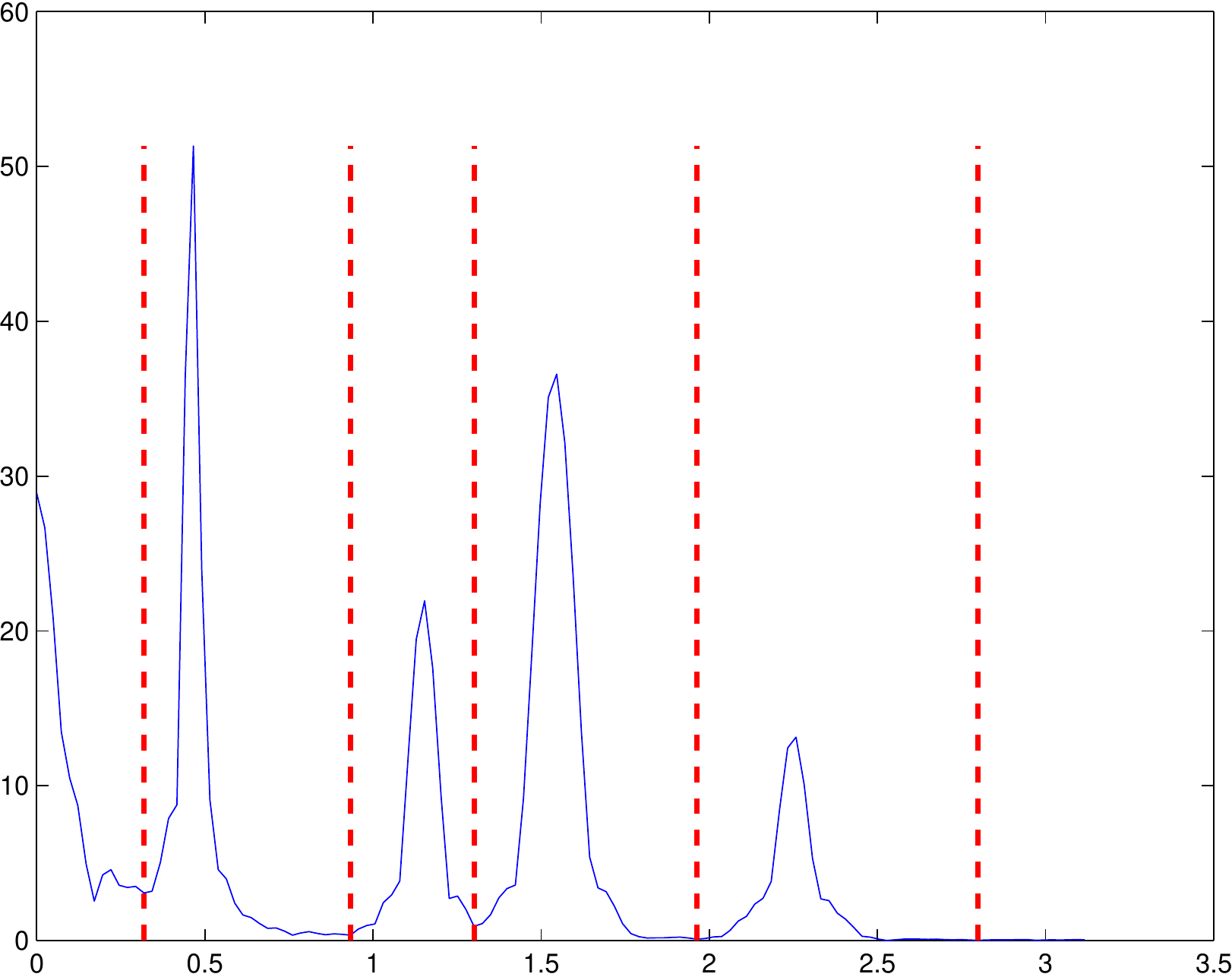} & \\
Otsu &
\end{tabular}
\caption{Boundaries for $Textures$.}
\label{fig:btextures}
\end{center}
\end{figure}

\begin{figure}[!h]
\begin{center}
\begin{tabular}{cc}
\includegraphics[width=0.47\textwidth]{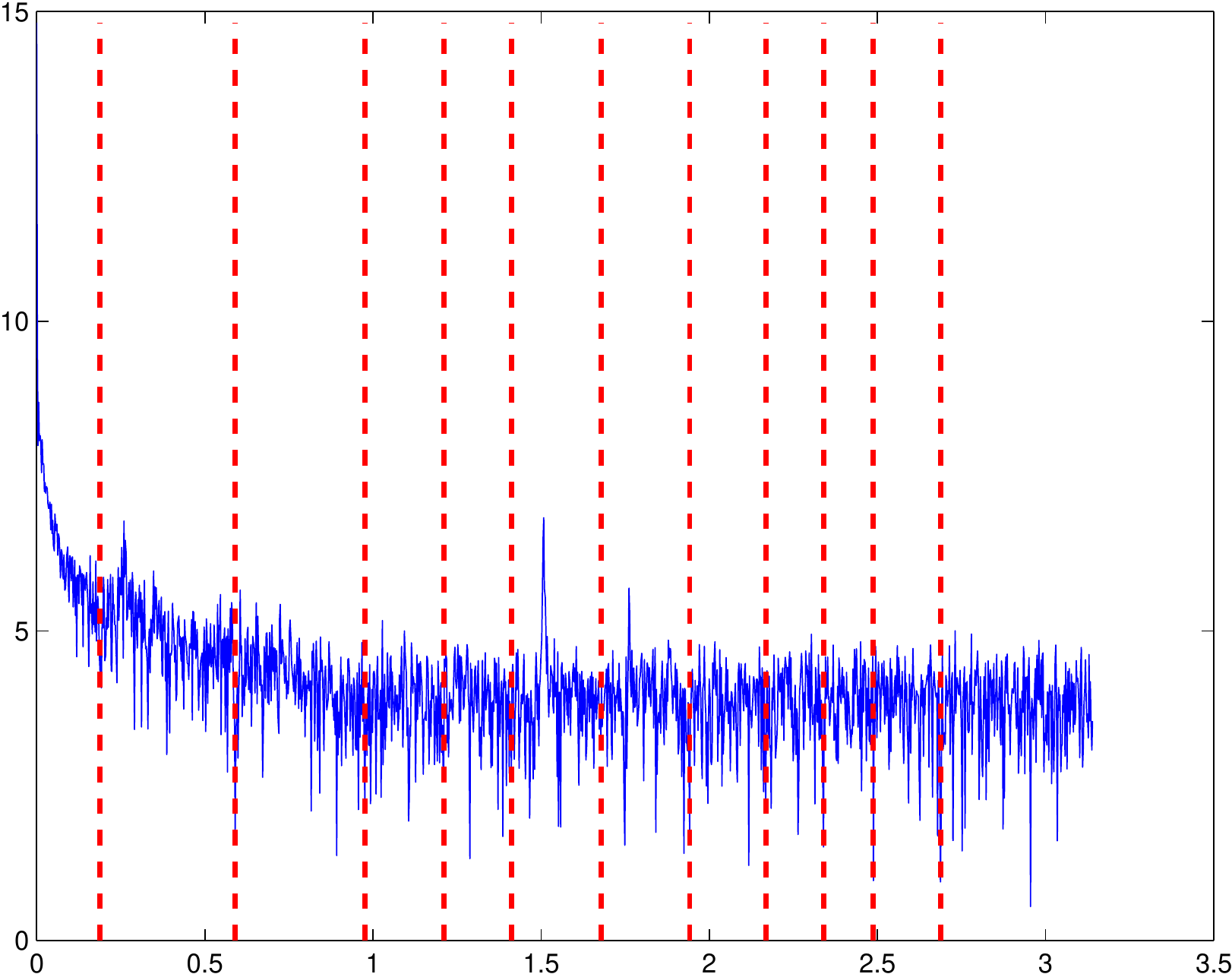} &
\includegraphics[width=0.47\textwidth]{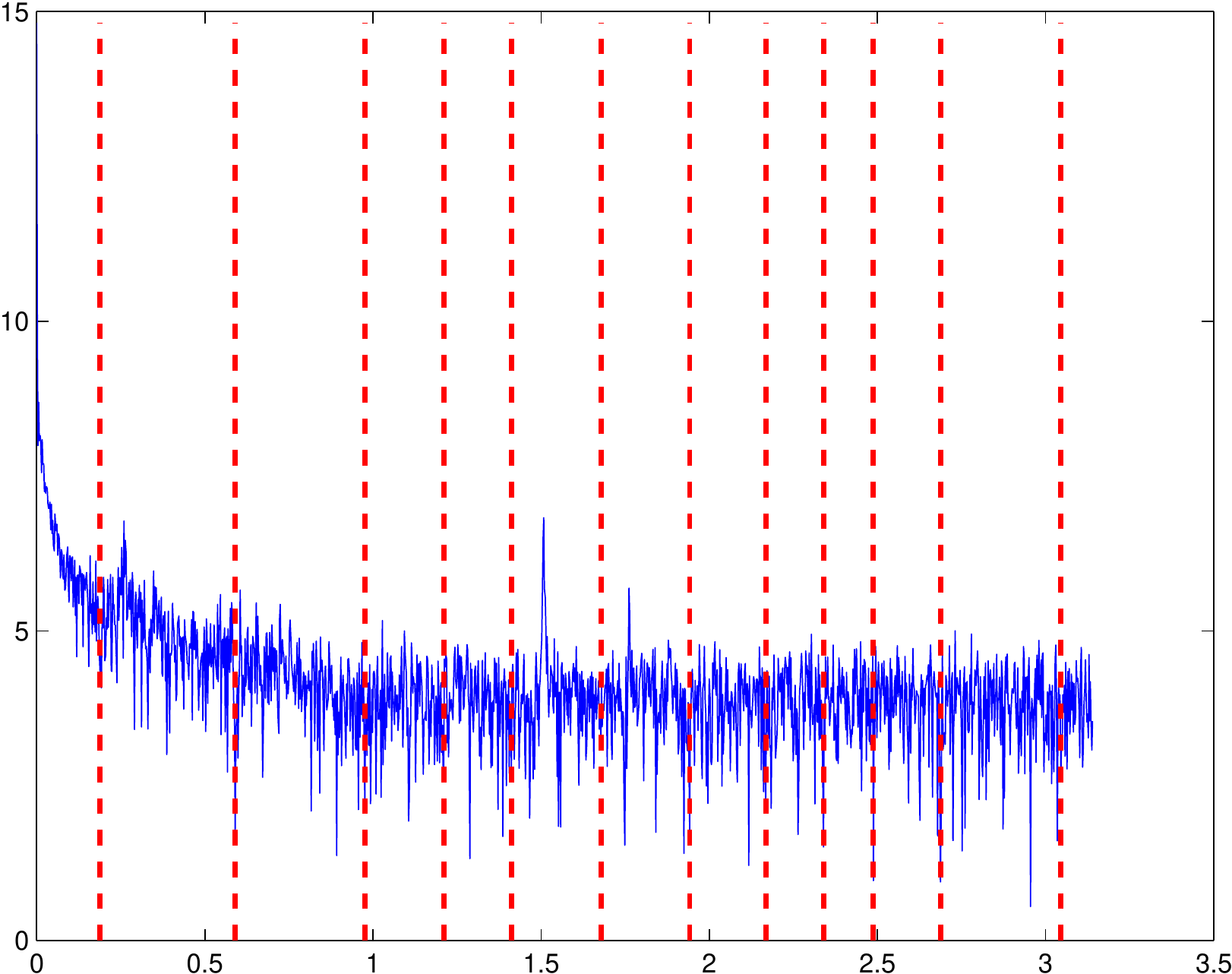} \\
$k-$Means & Otsu \\
\includegraphics[width=0.47\textwidth]{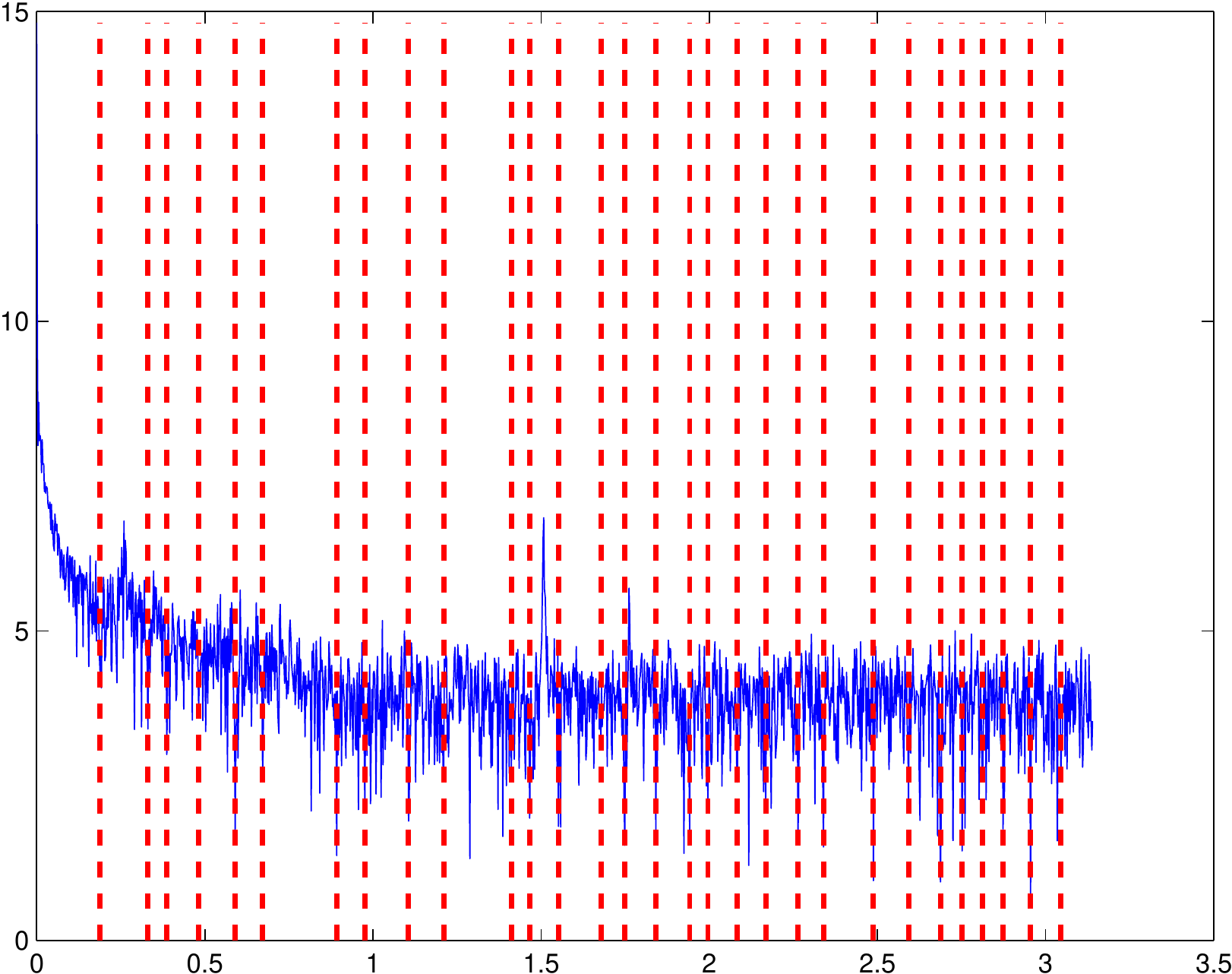} &
\includegraphics[width=0.47\textwidth]{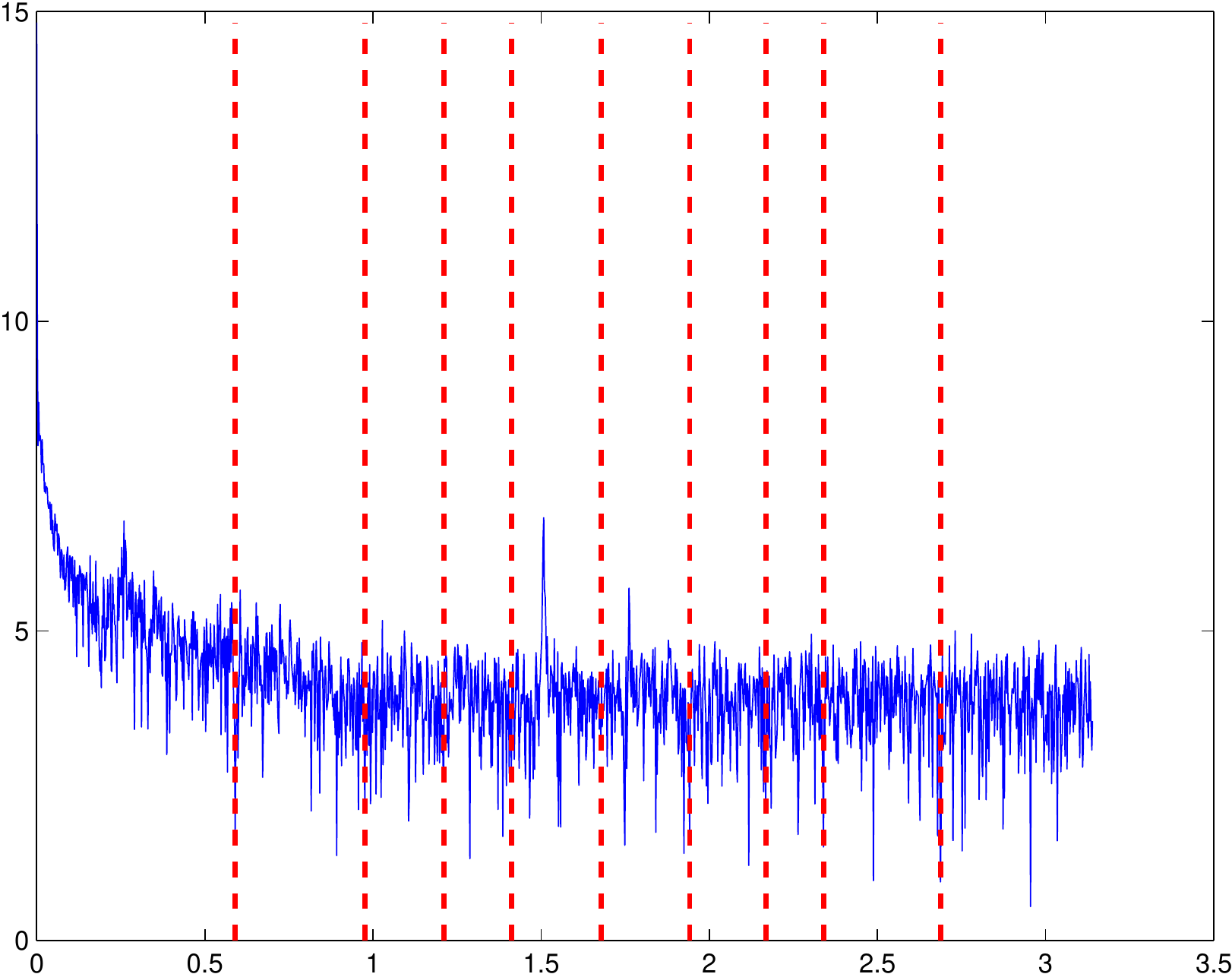} \\
Half-Normal & Empirical law \\
\end{tabular}
\caption{Boundaries for $EEG$.}
\label{fig:bsig6}
\end{center}
\end{figure}

In this section, we present the results obtained on 1D histograms by the method described in this paper. 
In figures~\ref{fig:hx16} and \ref{fig:hx21}, the histograms used (denoted $x16$ and $x21$, respectively) are histograms of 
grayscale values of two input images. Each of them contain 256 bins while in figures~\ref{fig:bsig1}, \ref{fig:bsig2}, 
\ref{fig:bsig3}, \ref{fig:bsig6} and \ref{fig:btextures} the used histograms are Fourier spectra, originally used in \cite{ewt1d} and 
\cite{ewt2d}. We call them $sig1,sig2,sig3,EEG$ and $Textures$, respectively. The obtained number of boundaries are presented in 
table~\ref{tab:bounds}.\\
First, note that the result corresponding to the empirical distribution is not 
shown for $x16$ because it does not provide relevant boundaries. Secondly, denote that, for all experiments, we do not provide the 
outputs from the uniform distribution. Indeed, in practice this distribution detects only the curves that have 
maximal lengths; it returns a single or a very few number of boundaries and misses important ones. It means that the uniform 
distribution assumption is not interesting for such detection problem. We can observe that for $x16$ and $x21$ all methods give 
acceptable sets of boundaries. For $sig1,sig2,sig3$ and $Textures$ spectra, Otsu's method and $\ell^2-k-$Means seem to provide 
the most consistent results throughout the different cases. Except for $x16$ and $Textures$, the half-normal distribution give similar 
results as Otsu's method. It can also be observed that the type of initialization (uniform or random) of the $k-$Means algorithm 
has no 
influence on the obtained boundaries. Moreover, except for the $Textures$ case, the $\ell^1-k-$Means and $\ell^2-k-$Means provide 
exactly the same results. Let us now comment the special case of $EEG$ spectrum. This spectrum is the spectrum of an electroencephalogram (EEG) 
signal and is much more complicated than usual spectra. As such it is difficult to have an a priori idea of a relevant number of modes 
as well as where they may occur. 
However, we can see that Otsu's method and $k-$Means give very similar outputs while the empirical distribution seems to reject 
extreme boundaries. Besides, the half-normal distribution generate a significantly higher number of modes.\\

{%
\newcommand{\mc}[3]{\multicolumn{#1}{#2}{#3}}
\begin{table}
\begin{center}
\begin{tabular}{l|ccccccc}
 & $x16$ & $x21$ & $Sig_1$ & $Sig_2$ & $Sig_3$ & EEG & Textures\\ \hline
$\ell^2-k-$Means (Random) & 6 & 2 & 1 & 2 & 3 & 11 & 4\\
$\ell^2-k-$Means (Uniform) & 6 & 2 & 1 & 2 & 3 & 11 & 4\\
$\ell^1-k-$Means (Random) & 6 & 2 & 1 & 2 & 3 & 11 & 3\\
$\ell^1-k-$Means (Uniform) & 6 & 2 & 1 & 2 & 3 & 11 & 3\\
Otsu & 6 & 3 & 2 & 3 & 4 & 12 & 5\\
Half-Normal law & 6 & 3 & 2 & 2 & 3 & 30 & 3\\
Empirical law & 1 & 2 & 1 & 2 & 1 & 8 & 2\\
\end{tabular}
\caption{Number of detected boundaries per signal for each detection method.}
\label{tab:bounds}
\end{center}
\end{table}
}%

\subsection{Grayscale image segmentation}
\begin{figure}[!t]
\begin{center}
\begin{tabular}{cc}
\includegraphics[width=0.47\textwidth]{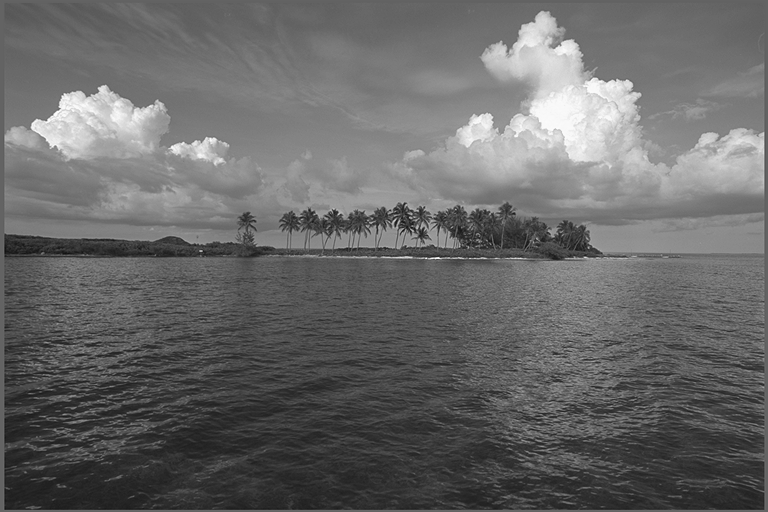} &
\includegraphics[width=0.47\textwidth]{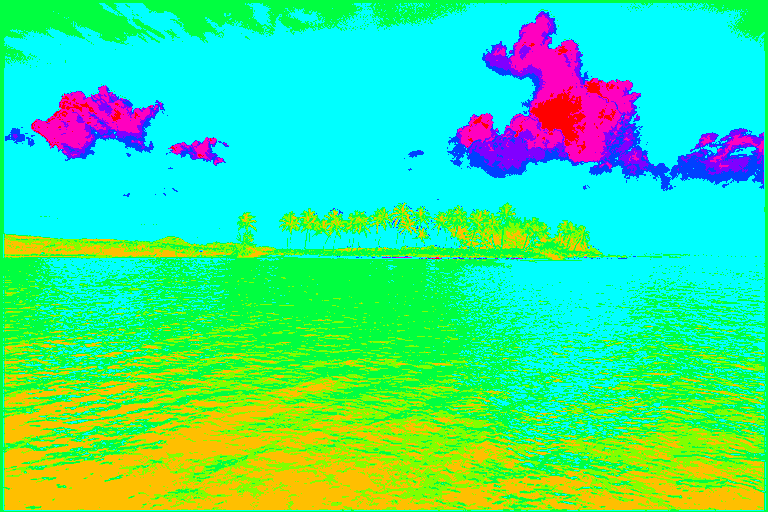} \\
Original & Segmented (8 classes)
\end{tabular}
\caption{Grayscale image segmentation: $x16$ case.}
\label{fig:x16}
\end{center}
\end{figure}

\begin{figure}[!h]
\begin{center}
\begin{tabular}{cc}
\includegraphics[width=0.47\textwidth]{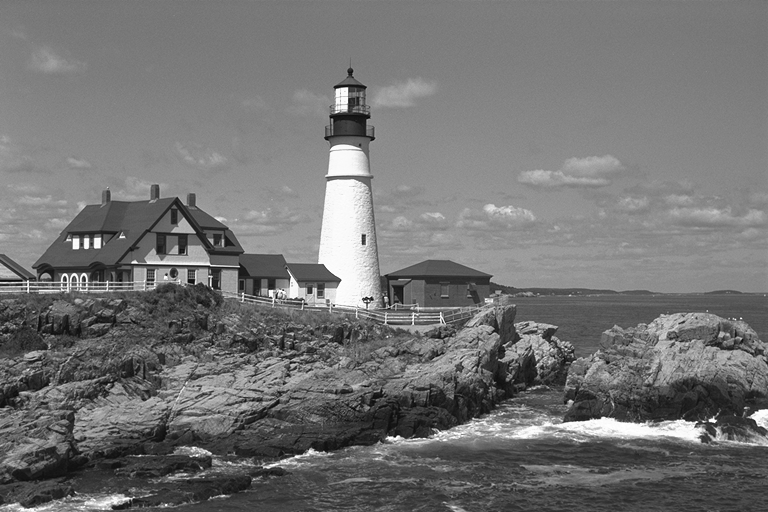} &
\includegraphics[width=0.47\textwidth]{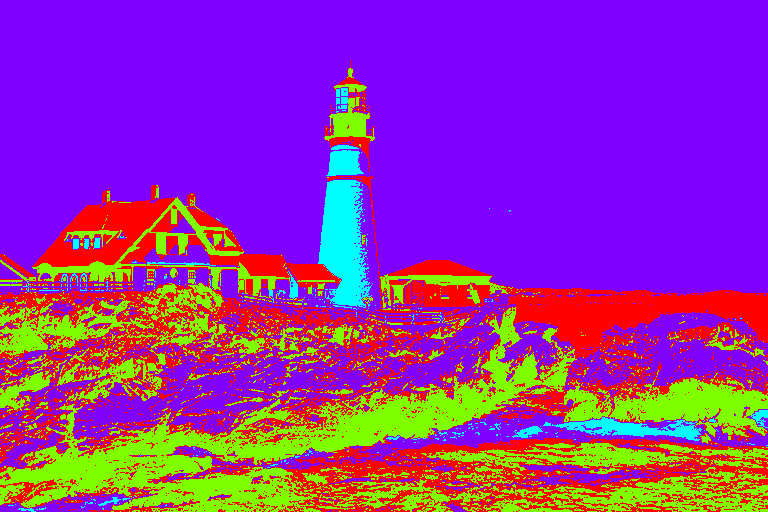} \\
Original & Segmented (4 classes)
\end{tabular}
\caption{Grayscale image segmentation: $x21$ case.}
\label{fig:x21}
\end{center}
\end{figure}

In this section, we address the grayscale image segmentation problem. This problem can be easily solved by our method; we segment the 
grayscale values histogram of the image which will automatically provide a certain number of classes.
Based on the previous section on 1D histograms, we choose to use Otsu's method in all following experiments (as well as in the 
next section). In figures~\ref{fig:x16} and \ref{fig:x21}, we present the images corresponding to the previous $x16$ and $x21$ histograms 
(the original image is on left and the segmented one on right). In both cases, this simple segmentation algorithm gives 
pretty good results as, for instance, it can separate important features in the images (clouds, sky, house's roof, light house, 
\ldots).

\subsection{Image color reduction}
\begin{figure}[!h]
\begin{center}
\begin{tabular}{cc}
\includegraphics[width=0.47\textwidth]{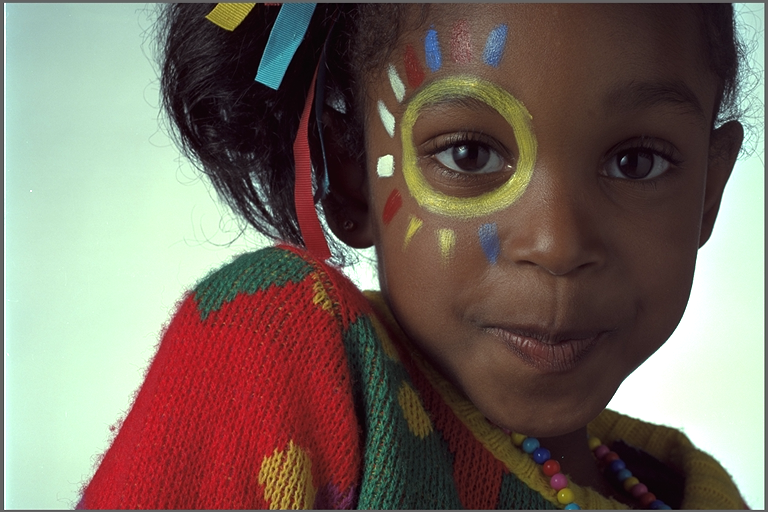} &
\includegraphics[width=0.47\textwidth]{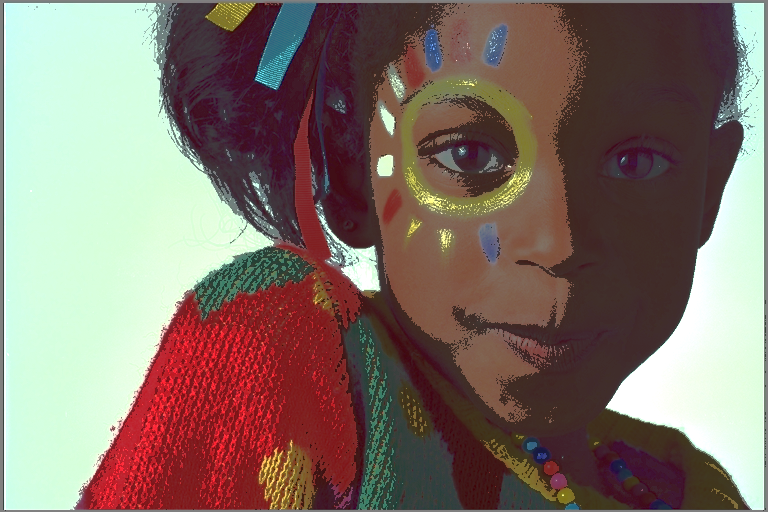} \\
Original & Reduced
\end{tabular}
\caption{Color reduction: $c15$ case.}
\label{fig:c15}
\end{center}
\end{figure}

\begin{figure}[!h]
\begin{center}
\begin{tabular}{cc}
\includegraphics[width=0.47\textwidth]{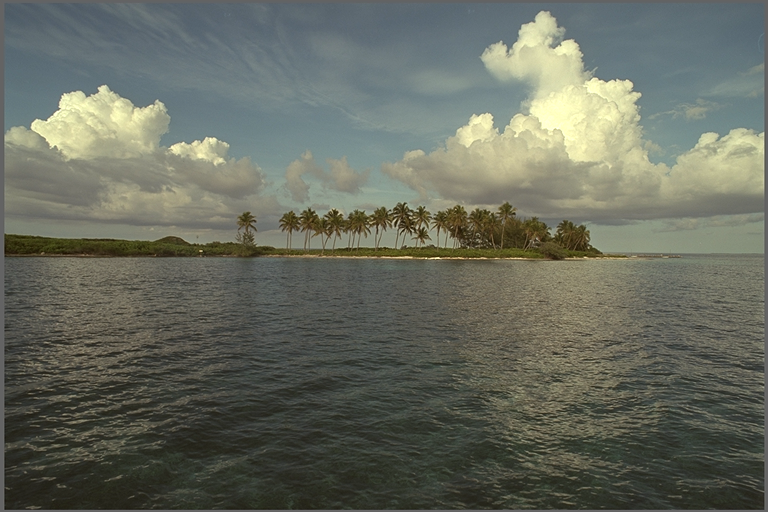} &
\includegraphics[width=0.47\textwidth]{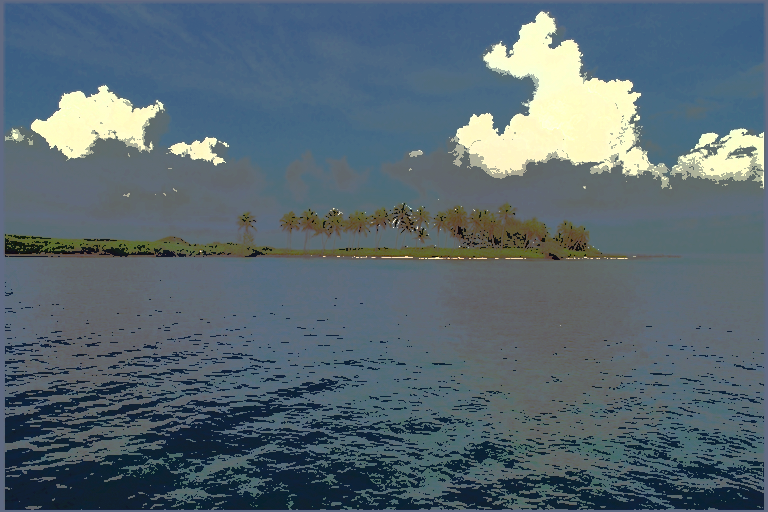} \\
Original & Reduced
\end{tabular}
\caption{Color reduction: $c16$ case.}
\label{fig:c16}
\end{center}
\end{figure}

\begin{figure}[!h]
\begin{center}
\begin{tabular}{cc}
\includegraphics[width=0.47\textwidth]{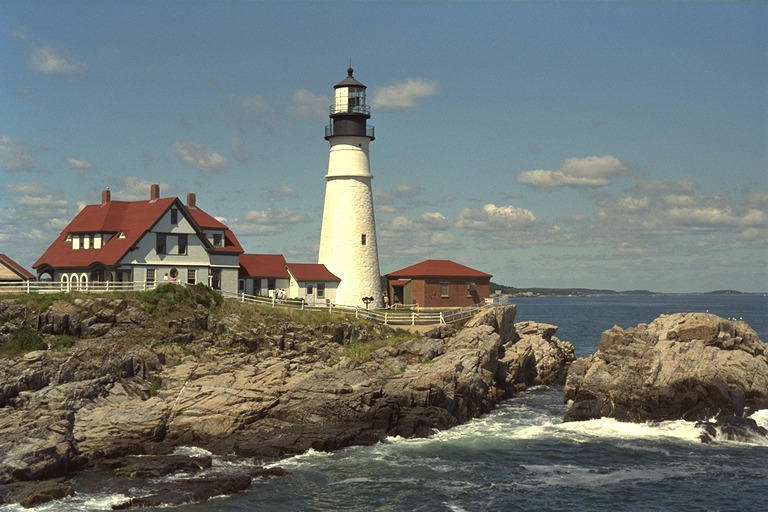} &
\includegraphics[width=0.47\textwidth]{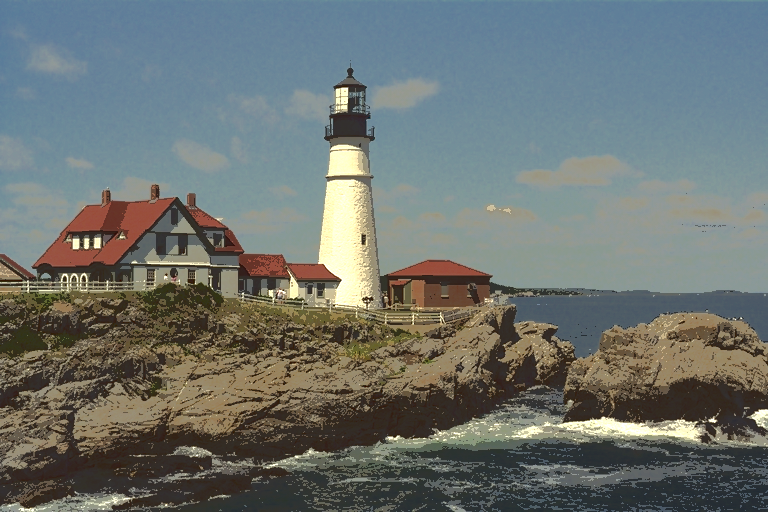} \\
Original & Reduced
\end{tabular}
\caption{Color reduction: $c21$ case.}
\label{fig:c21}
\end{center}
\end{figure}

\begin{figure}[!h]
\begin{center}
\begin{tabular}{cc}
\includegraphics[width=0.47\textwidth]{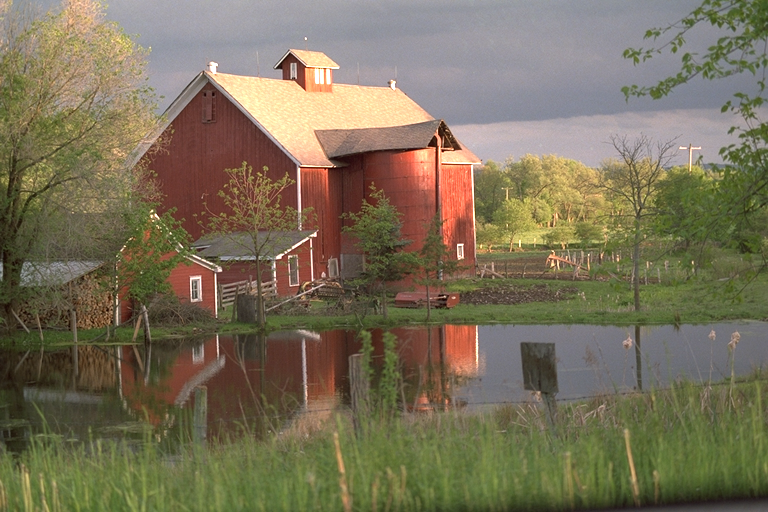} &
\includegraphics[width=0.47\textwidth]{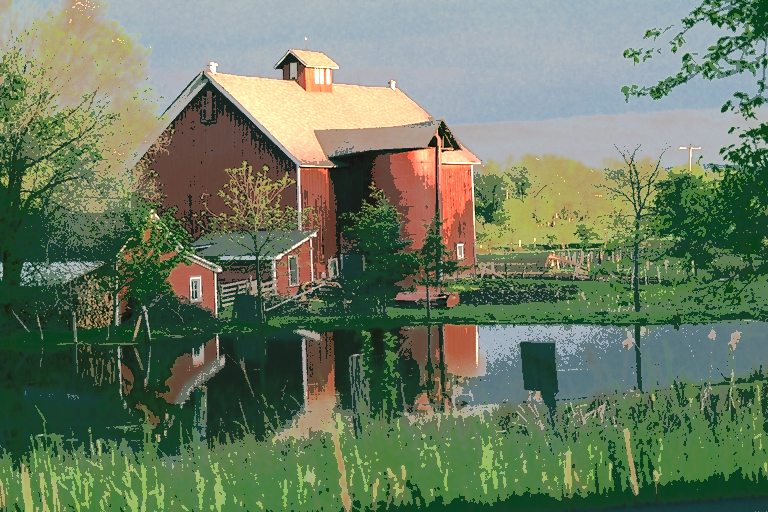} \\
Original & Reduced
\end{tabular}
\caption{Color reduction: $c22$ case.}
\label{fig:c22}
\end{center}
\end{figure}

In \cite{Delon2007a}, the authors use their histogram segmentation algorithm to reduce the number of colors in an image. 
Their method uses the following steps: first the image is converted into the HSV color system. Secondly, a first segmentation 
is obtained by segmenting the histogram of the V component. Thirdly, for each previous obtained class they segment the corresponding 
S's histograms. This step gives them a more refined set of color classes. Finally, they perform the same step, but on the 
H's histograms of each new class. The final set of color classes is provided as an initialization to a $k-$Means algorithm that performs 
the final color extraction. In practice, the HSV conversion and the segmentation of the V's histogram is sufficient to keep the most 
important colors in the image.\\
In figure~\ref{fig:c15}, \ref{fig:c16}, \ref{fig:c21} and \ref{fig:c22}, we show some example of such color reduction. 
We can see that this simple algorithm performs very well by reducing the number of colors used in the image but still retains the 
image's significant features.

\section{Conclusion}\label{sec:conc}
In this paper, we proposed a very simple and fast method to find meaningful modes in an histogram or a spectrum. The algorithm is 
based on the consistency of local minima in a scale-space representation. We show with several experiments that this method 
efficiently finds such modes. We also provide straightforward image segmentation and color reduction results.\\
In terms of future work, it will be interesting to characterize the behavior of the scale-space curves, i.e. when they disappear 
according to the histogram characteristics. These theoretical aspects are very challenging to tackle and will be addressed in 
an upcoming article. From the experimental point of view, we observed that the detected boundaries for EEG spectra look promising. 
Indeed, for the cases when we get around twelve modes, those modes look like refined version of the usual spectral bands that have been used by 
neuroscientists for a century. We plan to perform our algorithm on a large dataset of EEG signals to see if we get 
consistent modes with respect to several subjects or at different moments. Finally, it will be interesting to extend the proposed 
approach to find higher dimensional modes in larger dimension histograms or spectra.

\section{Acknowledgements}
This work is partially founded by the following grants NSF DMS-0914856, ONR N00014-08-1-119, ONR N00014-09-1-360, 
the UC Lab Fees Research and the Keck Foundation. All images are part of the Kodak dataset \cite{kodak}.

\end{document}